\documentclass[letterpaper, 10 pt, conference]{ieeeconf}

\IEEEoverridecommandlockouts                              
\usepackage{graphicx}
\usepackage{epstopdf} 		% for postscript graphics files
\usepackage{times} 			% assumes new font selection scheme installed
\usepackage{amsmath} 		% assumes amsmath package installed
\usepackage{amssymb}  		% assumes amsmath package installed
\usepackage{mathrsfs}
\usepackage{graphicx}
\usepackage{cases}
\usepackage[tight,footnotesize]{subfigure}
\usepackage{multirow}
\usepackage{enumerate}
\usepackage{color}
\usepackage{flushend}

\usepackage[utf8]{inputenc}
\usepackage{cleveref}
\crefname{section}{§}{§§}
\Crefname{section}{§}{§§}

\usepackage[T1]{fontenc}

\definecolor{light}{rgb}{0.8,0.8,0.8}
\definecolor{medium}{rgb}{0.6,0.6,0.6}
\definecolor{dark}{rgb}{0.4,0.4,0.4}
\definecolor{darkest}{rgb}{0.2,0.2,0.2}
\definecolor{Black}{rgb}{0,0,0}
\definecolor{White}{rgb}{1,1,1}
\definecolor{lightpurple}{rgb}{0.78823,0.709803,0.74509}
\definecolor{lightpurpletext}{rgb}{0.788235,0.5529411,0.658823}
\definecolor{skyblue}{rgb}{0.80392,0.866666,0.92941}
\definecolor{skybluetext}{rgb}{0.61568627,0.7647058,0.913725}
\definecolor{foliagegreen}{rgb}{0.3137254,0.458823,0.18431}
\definecolor{steelbluegrey}{rgb}{0.1961,0.2353,0.2392}

\usepackage{ctable}
\usepackage{longtable}
\newcolumntype{Y}{>{\centering\arraybackslash}X}
\newcolumntype{L}[1]{>{\raggedright\let\newline\\\arraybackslash\hspace{0pt}}m{#1}}
\newcolumntype{C}[1]{>{\centering\let\newline\\\arraybackslash\hspace{0pt}}m{#1}}
\newcolumntype{R}[1]{>{\raggedleft\let\newline\\\arraybackslash\hspace{0pt}}m{#1}}

\title{\LARGE \bf Assisted Control for Semi-Autonomous Power Infrastructure \\Inspection using Aerial Vehicles}

\author{Aaron McFadyen$^{1}$, Feras Dayoub$^{1,2}$, Steve Martin$^{1,2}$, Jason Ford$^{1,2}$ and Peter Corke$^{1,2}$% <-this % stops a space
\thanks{$^{1}$ is with the Science \& Engineering Faculty,
        Queensland University of Technology, Brisbane, Australia. {\tt\small (aaron.mcfadyen,j2.ford)@qut.edu.au}}
        \thanks{$^{2}$ is with the Australian Centre for Robotic Vision, Queensland University of Technology, Brisbane, Australia. {\tt\small (firstname.lastname)@qut.edu.au}}%, luis.mejias@qut.edu.au}}%
}

\begin{document}
\maketitle
\thispagestyle{empty}
\pagestyle{empty}

\begin{abstract}
This paper presents the design and implementation of an assisted control technology for a small multirotor platform for aerial inspection of fixed energy infrastructure. Sensor placement is supported by a theoretical analysis of expected sensor performance and constrained platform behaviour to speed up implementation. The optical sensors provide relative position information between the platform and the asset, which enables human operator inputs to be autonomously adjusted to ensure safe separation. The assisted control approach is designed to reduced operator workload during close proximity inspection tasks, with collision avoidance and safe separation managed autonomously. The energy infrastructure includes single vertical wooden poles and crossarm with attached overhead wires. Simulated and real experimental results are provided. 

\end{abstract}

\IEEEpeerreviewmaketitle

\section{Introduction}

The manual operation of unmanned aerial vehicles (UAV) for the purpose of close proximity inspection of infrastructure is becoming increasingly common. However, precise manual piloting of these platforms when undertaking inspections is a challenging task due to the pilot's difficulty in maintaining awareness and precise offset control of the UAV relative to the infrastructure. This requires highly skilled pilots to be located in close proximity to the asset. These aspects limit the efficiency and impedes potential value that could be realised from a UAV-based approach for infrastructure inspection. To address pilot limitations, attention is shifting towards robotic solutions mainly to provide the capability of obstacle avoidance and path planning. 

The majority of approaches to collision avoidance focus on full autonomy. However, in practice, an assisted autonomy solution is more palatable to regulators and also more desirable to end users. From the regulatory point of view, an assisted autonomous control system will have a human-in-the-loop which defines responsibilities and it is considered as a fail-safe against full autonomous control failure. From the end user point of view, it allows an expert operator to focus on hard-to-detect defects in the infrastructure without worrying about collisions.

\begin{figure}[t!]
\centerline{
\begin{tabular}{c}
{\includegraphics[angle=90,scale=0.07,trim = 5cm 2.5cm 5cm 3cm,clip=true]{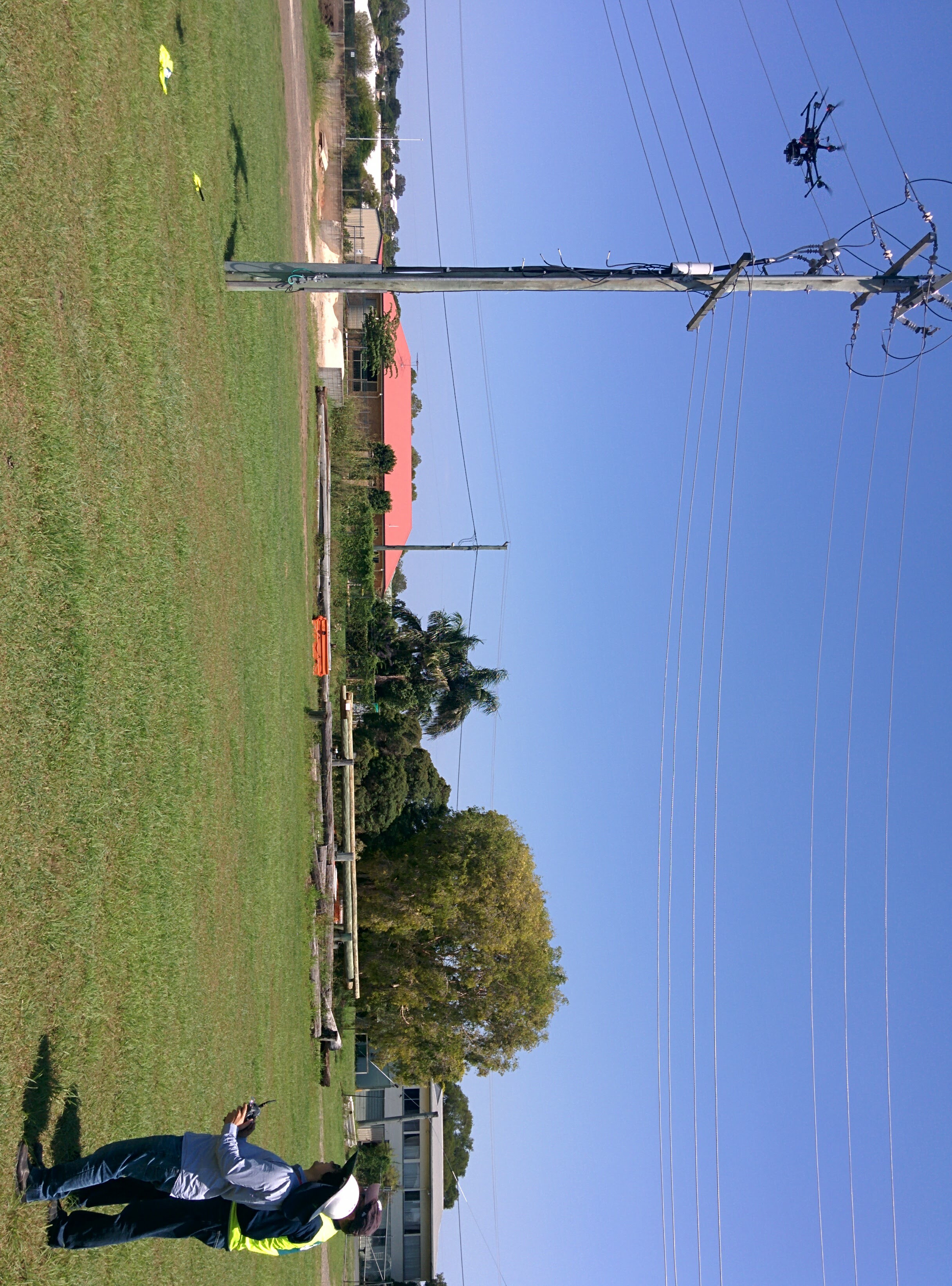}} 
\end{tabular}}
\caption[]{Example Flight trial showing the multirotor (DJIS800) and sensor rig (Intel NUC, Two RealSense cameras) inspecting an unenergized power pole, crossarm and wire configuration in Queensland, Australia.}
\label{exppic}
\end{figure}

This work addresses the task of inspecting electrical pole infrastructure using an unmanned aerial vehicle (UAV), Fig.~\ref{exppic}. The aim is to maintain a safety buffer zone to the electrical pole in order to increase the safety of the operation and improve the quality of collected data and also to reduce the pilot cognitive load and reduce the level of pilot skill required for close proximity pole inspection.

The contributions of this paper are:
\begin{enumerate}[$\blacktriangleright$]
\item A theoretical analysis of expected sensor performance and constrained platform behaviour.
\item A light weight vision only, computationally efficient approach to electrical pole detection and avoidance. 
\end{enumerate} 

This paper is structured as follows. Section \ref{back} provides some background on UAVs for autonomous inspection. Section \ref{prelims} presents the operational concept for assisted inspection of power infrastructure. Section \ref{pj} provides an overview of the hardware and software selection for perception.  ection \ref{cont} describes the assisted control approach whilst Section \ref{sec:exp} provides simulated and real experimental results for a multirotor inspecting a fixed power pole. Section \ref{conc} offers concluding remarks.

\section{Background} 
\label{back}

The majority of work on infrastructure inspection focus on the autonomous aspect of the problem, as shown in recent survey by M{\'a}th{\'e} et al.~\cite{Mth2015VisionAC}. The assumption is that the data recorded during the autonomous operation can be processed offline by experts as in the case of railway semaphore inspection~\cite{Mth2016VisionbasedCO}, wind turbine inspection \cite{Stokkeland2015AutonomousVN} and power line inspection~\cite{Li2008KnowledgebasedPL}. However, an autonomous inspection requires the use of an expert system onboard the UAV that replaces the human expert and can steer the platform towards potentially defective parts of the asset. This is possible in the case of simple structures, such as power lines \cite{Araar2014VisualSO}, but not always possible when the structure is complex or the type of defect is unknown. In these cases a human expert is needed to steer the platform.% while provided with assisted autonomy.

Assisted autonomy of UAV focuses on reducing the possibility of collisions during missions. This can be done in a passive or an active way. Passively, the assistance is performed by directing the pilot using external forces through a haptic control device as demonstrated in \cite{Lam2009ArtificialFF,Brandt2010HapticCA,Hou2013RepresentationOV}. 
Whereas the active approach works by altering the direction and magnitude of the operators input velocity~\cite{Sa2015InspectionOP}, usually to ensure collision free motion when operating in close proximity to obstacles (e.g power infrastructure in the case of this work). Typically, the motivation behind this is so that the UAV maintains smooth and continuous motion during operation leading to improved data collection. 

However, in an assisted inspection mission, the operator often wants to position the platform such that a particular view of the infrastructure can be examined clearly, without accidentally causing a collision. In this case, it is undesirable to alter the direction of the controller input, instead we simply modify the magnitude such that the robot never breaches a predefined inner collision boundary. This paper proposes an approach to achieve this type of assisted autonomy for infrastructure inspection.

\section{Preliminaries} \label{prelims}

\subsection{Notation} \label{n}
A position vector is denoted by a single point $\mathbf{p}(x,y,z)$, where $x$, $y$ and $z$ are the scalar displacements of the point from an earth fixed reference coordinate frame $\mathcal{F}_R$ with origin $\mathcal{O}_R$. A tilde is used to denote a requested control input from the operator such that $\tilde{\mathbf{u}}$ and $\tilde{u}$ represent vector and scalar control inputs respectively. Unless otherwise stated, SI units are assumed.

The bounding polygon $\partial\mathcal{S}$ of a finite point set $\mathcal{S} \in \mathbb{R}^2$ is the set of points in the closure of $\mathcal{S}$, not belonging to the interior of $\mathcal{S}$. The bounding polyhedra $\partial\mathcal{V}$ of a finite point set $\mathcal{V} \in \mathbb{R}^3$ is the set of points in the closure of $\mathcal{V}$, not belonging to the interior of $\mathcal{V}$. 
\begin{figure}[t!]
\vspace{0pt}
\centerline{
\begin{tabular}{c}
{\includegraphics[scale=0.6,trim = 5cm 2.8cm 5cm 3cm,clip=true]{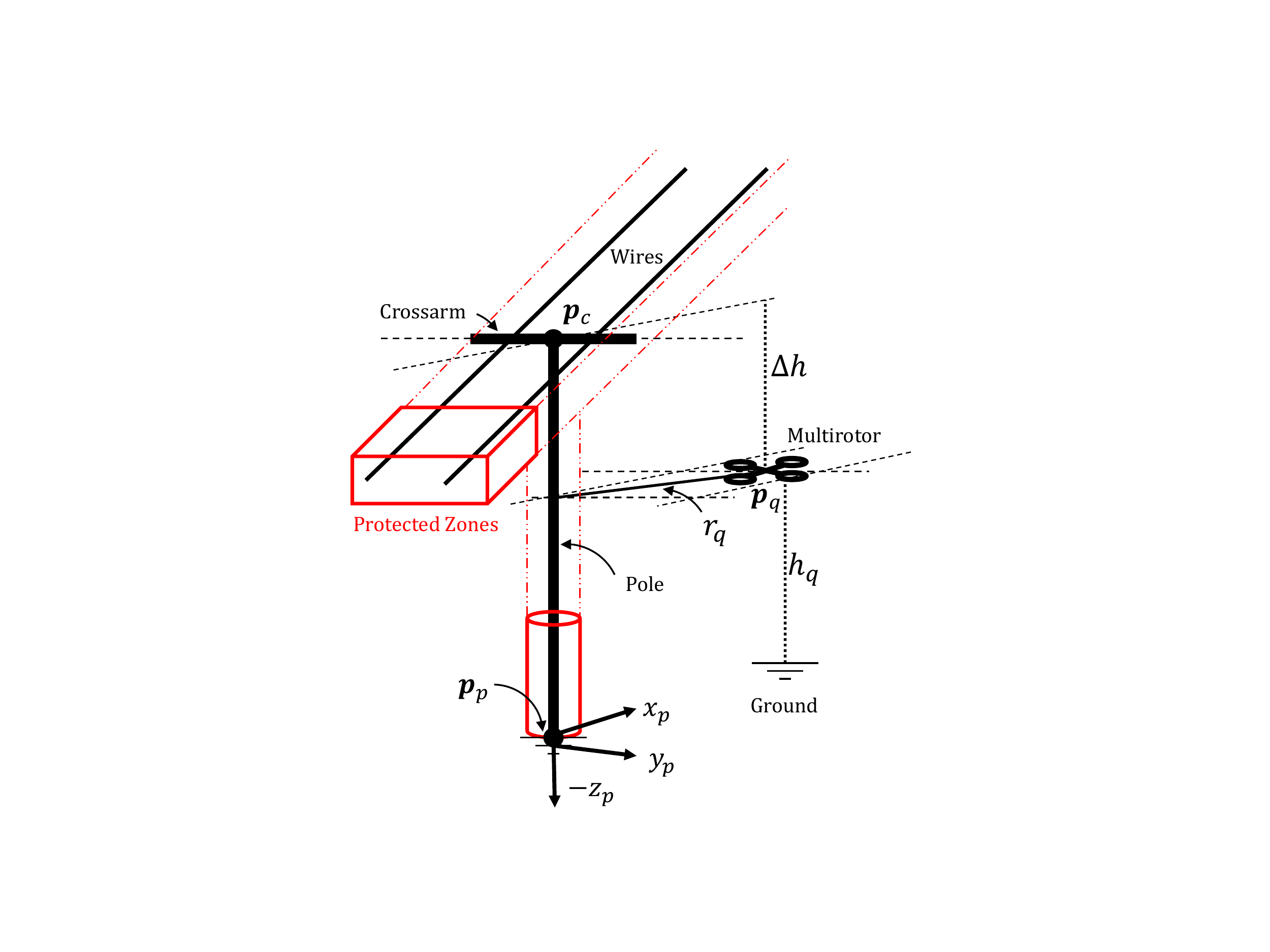}} 
\end{tabular}}
\caption[]{3D pole, crossarm and wire (\textcolor{black}{$\blacksquare$}) geometry and parameters including example protection zones (\textcolor{red}{$\square$}) and arbitrarily located platform. }
\vspace{-14pt}
\label{d1}
\end{figure}
\subsection{Infrastructure Parameters} \label{ia}
\begin{figure}[t!]
\vspace{0pt}
\centerline{
\begin{tabular}{c}
{\includegraphics[scale=0.5,trim = 5cm 2.8cm 5cm 3cm,clip=true]{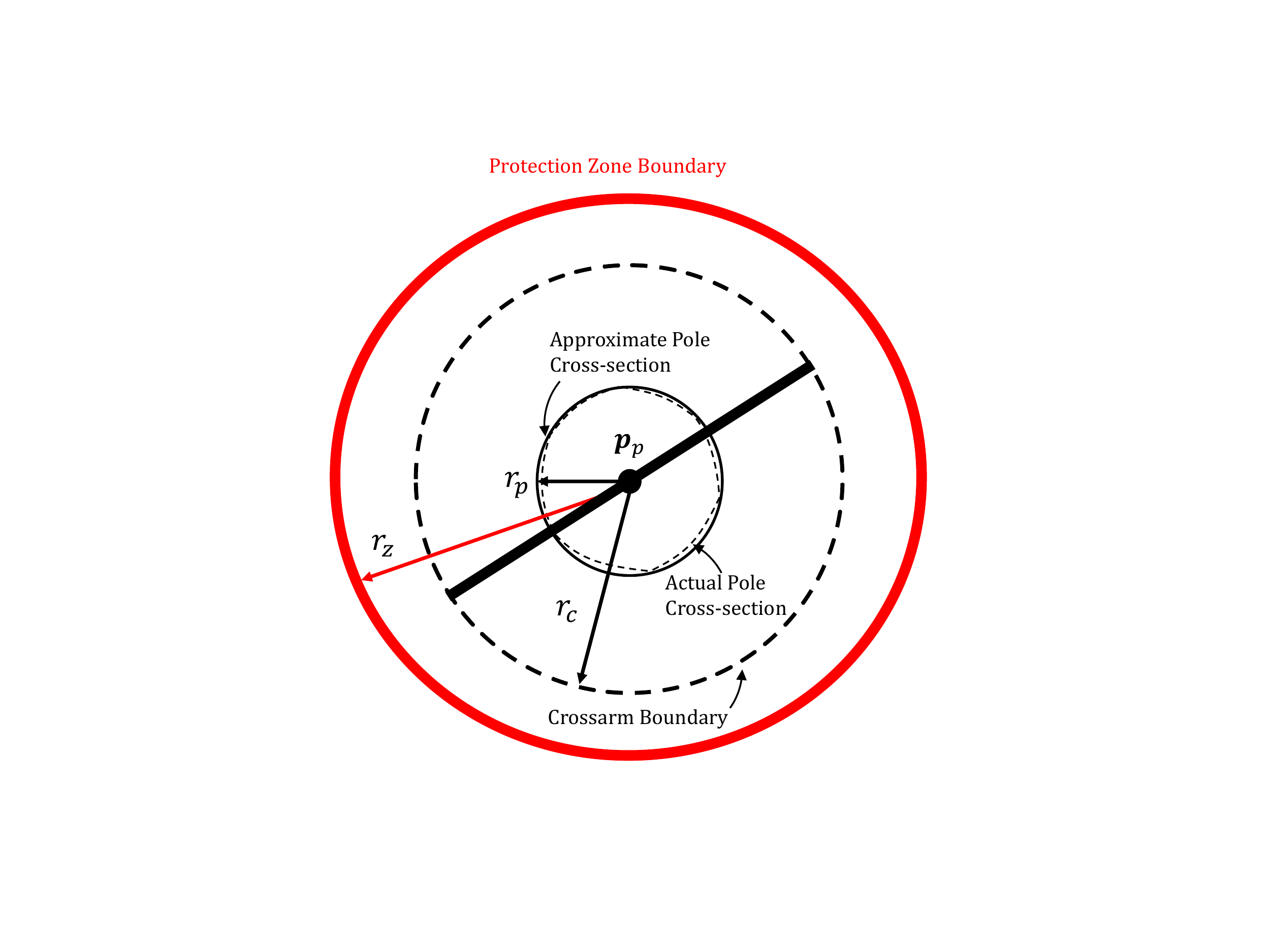}} 
\end{tabular}}
\caption[]{2D ($xy$-planar view) pole, crossarm and wire parameters including example protection zones  (\textcolor{red}{$\square$}) and crossarm region (\textcolor{black}{$--$}). }
\vspace{-14pt}
\label{d2}
\end{figure}

\noindent Assumptions regarding the infrastructure are as follows:
\begin{enumerate}[1.]
\item Each pole has a height $h_p$ and radius $r_p$ such that {$6 \leq h_p \leq 10$ and $0.2 \leq r_p \leq 0.4$}. If a pole is not cylindrical, it can be modelled as a cylinder. The approximating cylinder encloses all edges of the pole and provides an upper bound of the pole size. Each pole is located at the position $\mathbf{p}_p$ which denotes the point where the centreline of the approximating cylinder intersects the ground plane. Pole tilt is no more than {$5^{\circ}$} from the $z$ axis in $\mathcal{F}_R$.
\item Each pole has a single cross arm attached to the top of the pole at point $\mathbf{p}_c$ such that the centre of the crossarm and the pole coincide. The crossarm has a radius $r_c$ measured from $\mathbf{p}_c$ to either end of the crossarm such that {$0.5 \leq r_c \leq 2$}. The crossarm resides at height $h_p$ and has an orientation $\psi_c \in (0, 2\pi)$ measured from the $x$ axis of $\mathcal{F}_R$. The crossarm is assumed to be perpendicular to the pole with unknown orientation.
\item Each pole has an circular crossarm region $\partial \mathcal{S}$ centred on $\mathbf{p}_c$ with radius $r_c$ where the crossarm may be located. Crossarm $\partial \mathcal{S}_c$ and pole $\partial \mathcal{S}_p$ detection regions are defined as the intersection of the sensor field of view and the circular crossarm region $\partial \mathcal{S}$.
\item Incoming and outgoing linear wires $w_i$ where $i \in \{1,\dots,4\}$ are attached to the crossarm. The wires run parallel to each other such that $w_i \parallel w_{j}$ for $i,j \in \{1,\dots,4\}$ and perpendicular to the crossarm such that $w_i \perp r_c$. Incoming wires have the same outgoing direction and are located in the same plane as $\partial \mathcal{S}_c$ and $\partial \mathcal{S}_p$.
\item Each pole, crossarm and wire configuration has a restricted volume or protection zone $\partial \mathcal{V}_z$. The protected zone is made up from two convex polyhedra consisting of a regular cylinder with radius $r_z$ centred on the pole and a rectangular polyhedron of infinite length, height $h_z$ and width $d_z$. The resulting non-convex polyhedron encloses the pole, crossarm and wires such that a vertical cross section resembles the letter `T' (see Fig.~\ref{d1} and \ref{d2}). 
\end{enumerate}

\subsection{Platform Parameters} \label{pa}
\noindent Assumptions regarding the platform are as follows:
\begin{enumerate}[1.]
\item The platform is a modified 6 rotor DJI S800 platform with maximum takeoff weight (MTOW) less than 10kg including payload. The platform payload consists of a computer (Intel NUC), application sensors (RealSense cameras), autopilot (including GPS receiver, IMU and barometer) and all supporting telemetry and control links.
\item The platform position $\mathbf{p}_q(x,y,z)$ defines the location of the platform centre of mass in $\mathcal{F}_R$. The platform position relative to the infrastructure is defined by the horizontal distance $r_q$ to the pole (or projection of the pole centreline to $\pm\, \infty$) and the vertical distance $\Delta h$ from the $xy$-plane located at the same altitude as the crossarm where $\Delta h < 0$ is below and $\Delta h > 0$ is above the crossarm. 
\item The platform is controlled using velocity commands in the platforms $x$, $y$, $z$ and $\psi$ (yaw) directions. The platform axis assumes an NED system, whereby the $x$ axis runs along the platform arm that intersects the FOV of both onboard camera sensors (see Remark 1 above).  The platform $z$ axis points down from the platform centre of mass such that its direction is aligned with the gravity vector in stable hover only.
\end{enumerate}

\section{Perception}\label{pj}
To ensure adequate perception for the assisted control, both hardware and software aspects are tackled. The hardware aspect involve sensor choice and placement whilst the software aspect consider algorithms for detection and data extraction.
\begin{figure}[t!]
\vspace{0pt}
\centerline{
\begin{tabular}{c}
{\includegraphics[scale=0.6,trim = 3cm 9cm 3cm 9cm,clip=true]{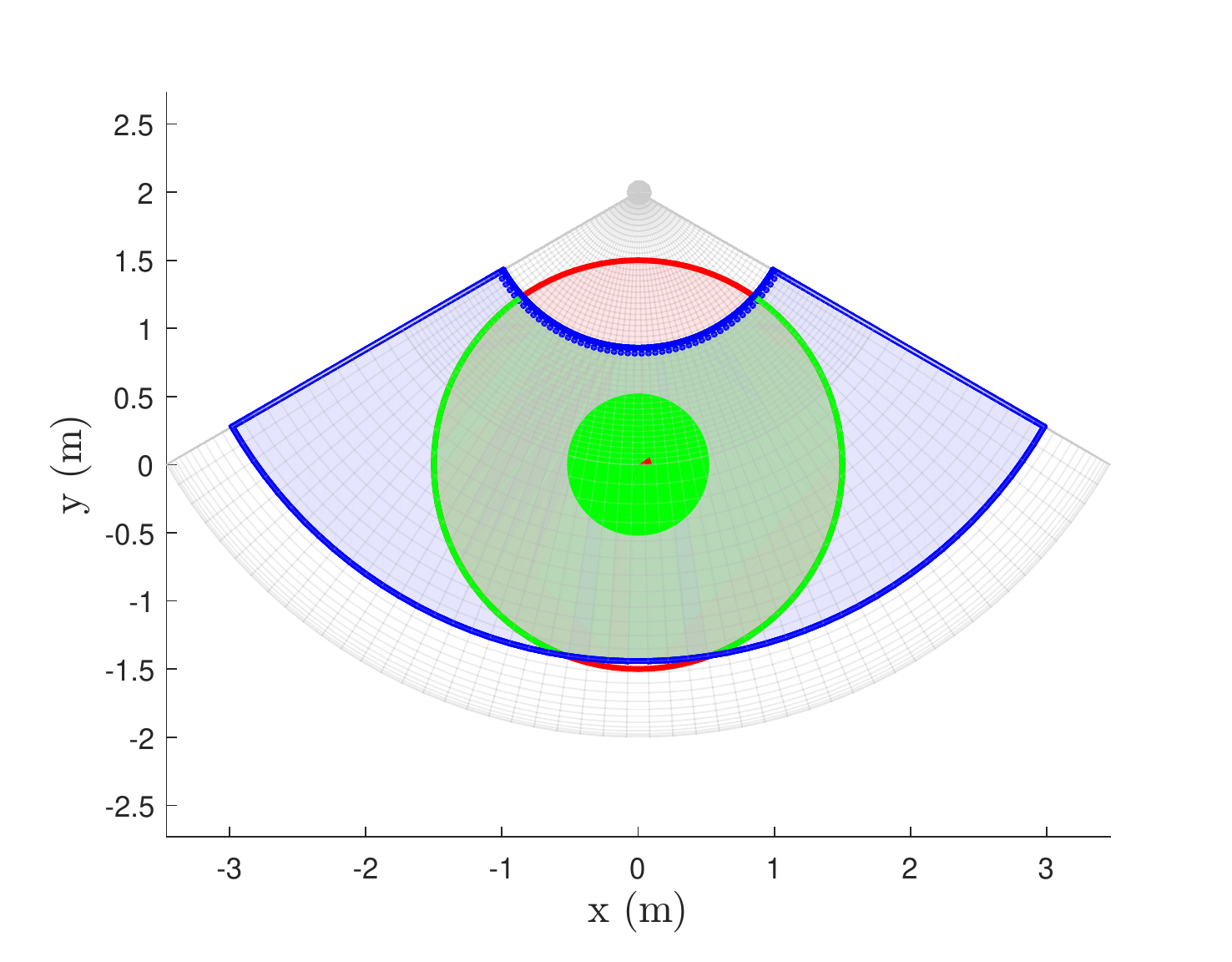}}  
\end{tabular}}
\vspace{-8pt}
\caption[]{Example crossarm region coverage for $\mathbf{p}_q(2,0,4)$, $\phi=\theta=\psi = 0$ and $\mathbf{u}=\mathbf{0}$. The RealSense field of view (\textcolor{gray}{$\square$}) using $H_C,V_C,R_C$ and crossarm region $\partial \mathcal{C}$  (\textcolor{red}{$\square$}) are shown along with the sensor coverage of the $xy$-plane at pole altitude $\partial \mathcal{S}_{xy}$ (\textcolor{blue}{$\square$}), sensor coverage of the crossarm region $\partial \mathcal{S}_{c}$ (\textcolor{green}{$\square$}) and sensor coverage of the pole region $\partial \mathcal{S}_{d}$  (\textcolor{green}{$\blacksquare$}).}
\vspace{-12pt}
\label{p1}
\end{figure}

\subsection{Sensor Configuration}\label{sensorConfig}
Sensor choice and placement are important considerations for inspection tasks. The location of the sensors should be such that the probability of detecting the target object is maximised, regardless of the platform orientation and before the application of any additional processing. The analysis conducted in this section informs the sensor choice and placement on the platform, by considering a theoretical treatment of the expected operation and associated constraints. The first analysis evaluates the crossarm detection probability using typical sensor field of view constraints to derive suitable protection zone boundaries. The second analysis uses a modified detection probability metric to enhance the inspection performance through flight control limitations and sensor placement parameters. %imposed by sensor and platform performance limits.. 

\begin{figure}[t!]
\vspace{0pt}
\centerline{
\begin{tabular}{c}
{\includegraphics[scale=0.5,trim = 3cm 9cm 3cm 9cm,clip=true]{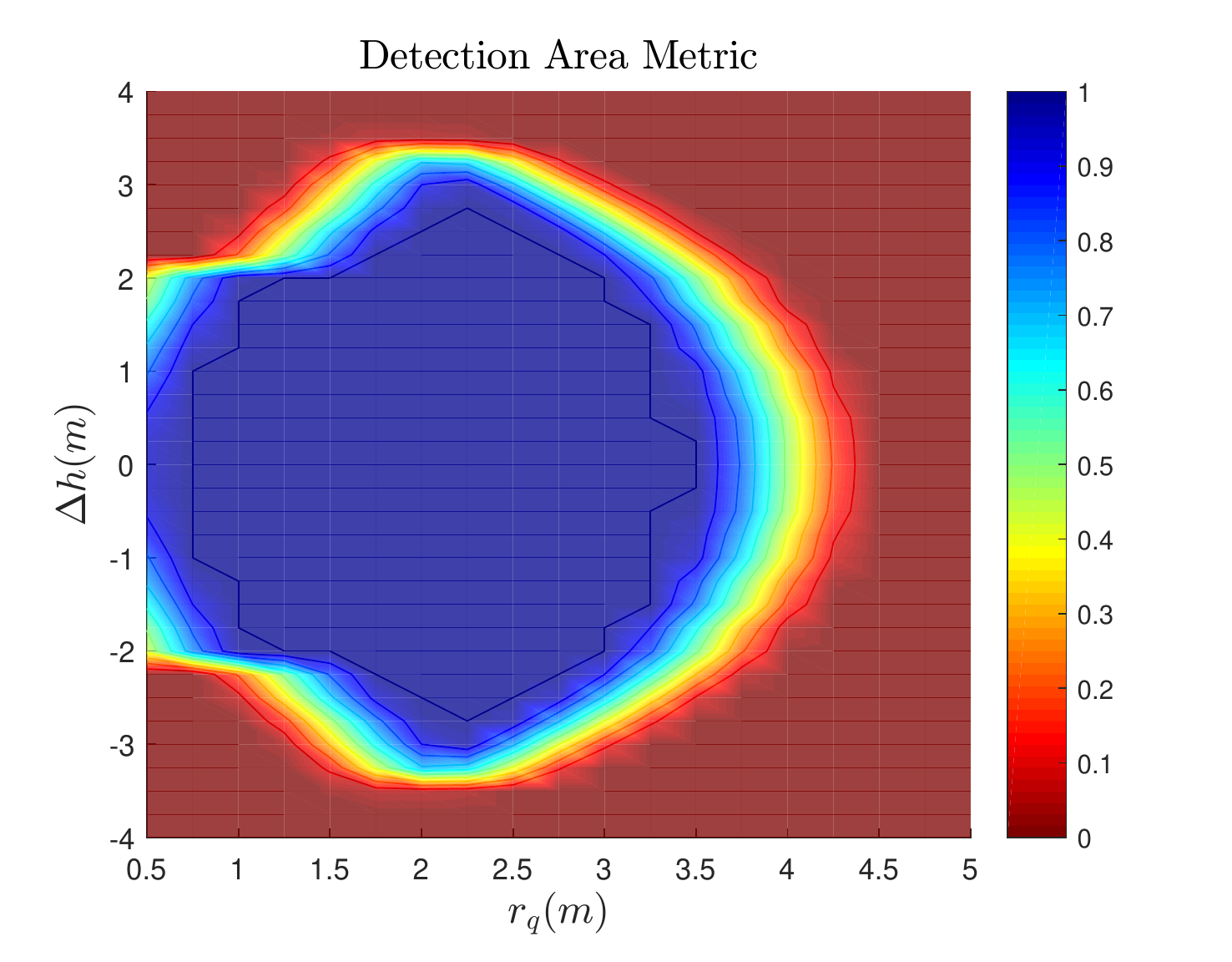}} 
\end{tabular}}
\vspace{-8pt}
\caption[]{Detection area metric $A_s/A_p$ with $\mathbf{p}_c(0, 0, 6)$, $r_c = 1.5$, $r_d=0.5$ for $r_q \in (0.5,5.0)$ and $\Delta h \in (-4,4)$. As the detection area metric goes from zero (\textcolor{red}{$\blacksquare$}) to unity (\textcolor{blue}{$\blacksquare$}), the sensor covers more of the pole region such that the probability of detected the crossarm increases, regardless of the crossarm orientation $\psi_c$.}
\vspace{-14pt}
\label{r1}
\end{figure}

\begin{figure}[t!]
\vspace{0pt}
\centerline{
\begin{tabular}{c}
{\includegraphics[scale=0.5,trim = 3cm 9cm 3cm 9cm,clip=true]{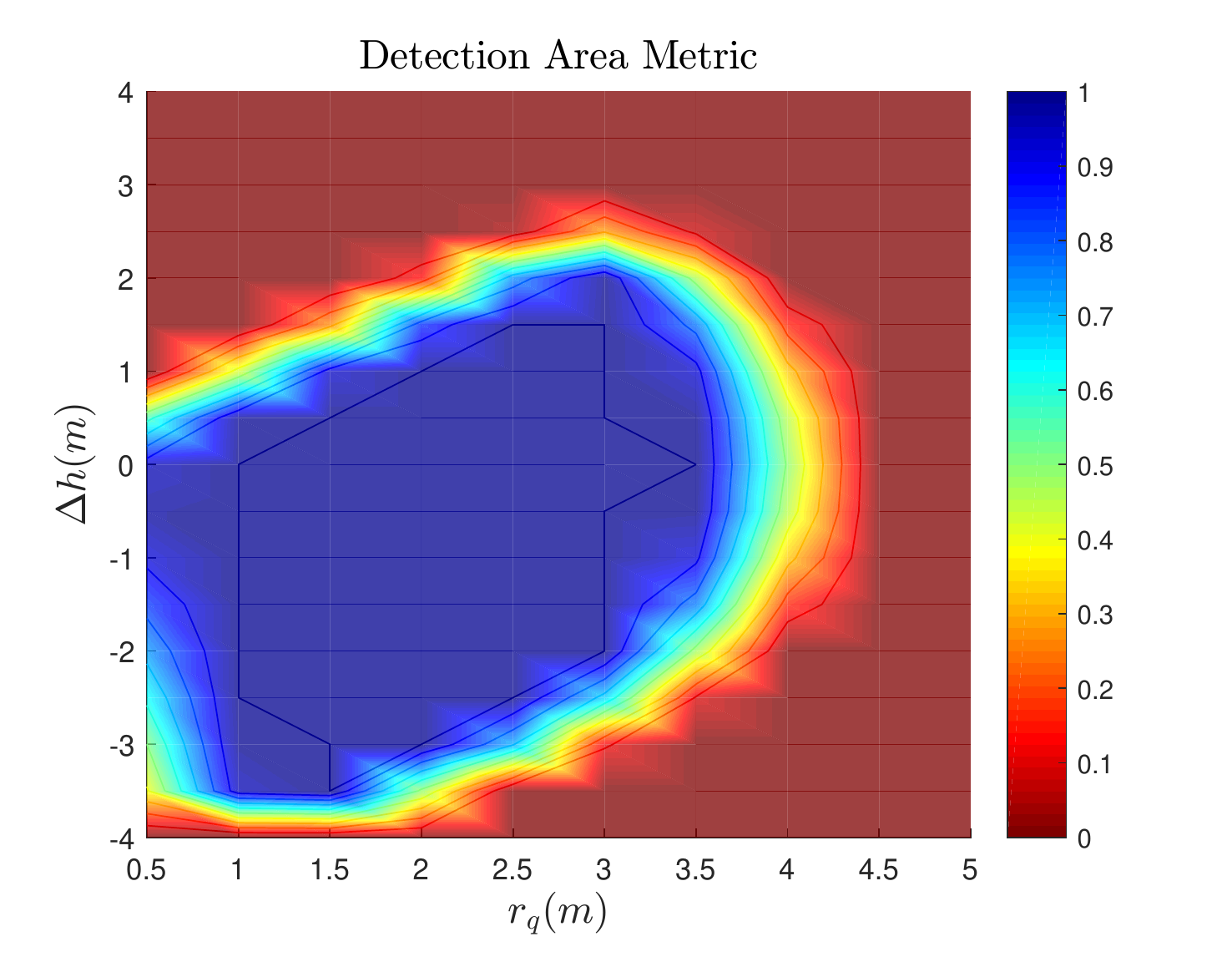}} 
\end{tabular}}
\vspace{-8pt}
\caption[]{Detection area metric $A_s/A_d$ with $\mathbf{p}_c(0, 0, 6)$, $r_c = 1.5$, $r_d=0.5$ for $r_q \in (0.5,5.0)$, $\Delta h \in (-4,4)$ and $\theta=20^{\circ}$. Notice the lack of coverage and reduced probability of detecting the crossarm when pitching up (moving away from the crossarm) and operating near the pole.}
\vspace{-14pt}
\label{r2}
\end{figure}

For the analysis presented in this section, the following inspection scenario is used. Consider a pole located at $\mathbf{p}_p(0,0,0)$ with $h_p = 6$ and $r_p = 0.2$. The crossarm and wire configuration is such that $\mathbf{p}_c(0, 0, 6)$ with $r_c = 1.5$ and $r_d=0.5$. The crossarm and wire orientation $\psi_c \in (0, 2\pi)$ is unknown. Consider also a platform located at $\mathbf{p}_q(x, y, z)$ with orientation $\phi = 0$, $\theta = 0$ and $\psi = 0$. The optical sensor(s) are rigidly attached to the platform such that the origin of the sensor(s) coincide with the platform centre of mass. The platform moves in the $x$ and $z$ direction such that $r_q \in (0.5,5.0)$ and $\Delta h \in (-4,4)$ with $y=0$. At each location, the platform may take a maximum (or minimum) control input corresponding to $(\phi=\pm 0^{\circ}, \theta=\pm 20^{\circ}, \dot{\psi}=\pm 30^{\circ/\text{s}}, \dot{v}_z=\pm 1 \text{m/s})$.  The sensor(s) field of view {$\{H_C,V_C\}$ is such that $-60 \leq H_C \leq 60$ degrees horizontally and  $-60 \leq V_C \leq 60$ degrees vertically with a detection range of $R_C=4$.}
\begin{figure}[t!]
\vspace{0pt}
\centerline{
\begin{tabular}{cc}
{\includegraphics[scale=0.21,trim = 0cm 0cm 0cm 0cm,clip=true]{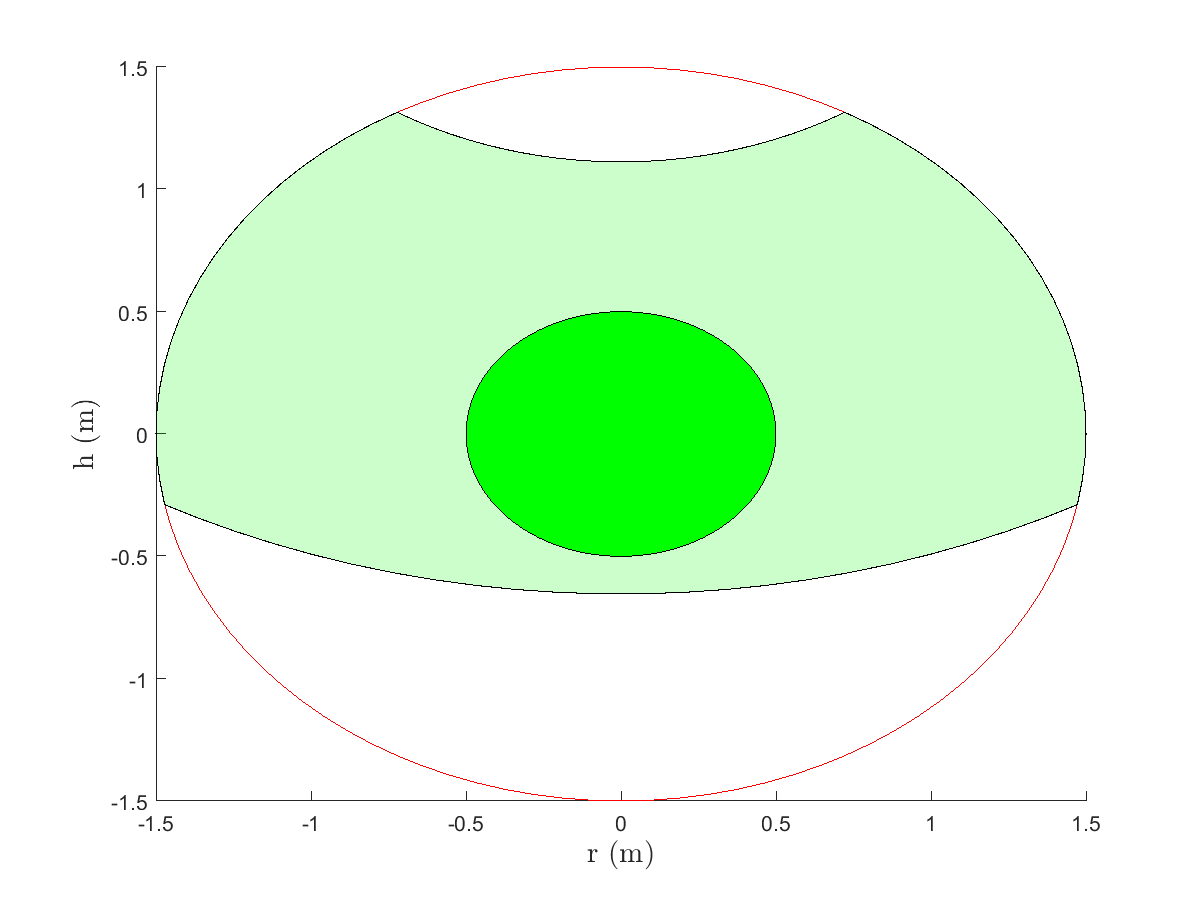}}  &
{\includegraphics[scale=0.21,trim = 0cm 0cm 0cm 0cm,clip=true]{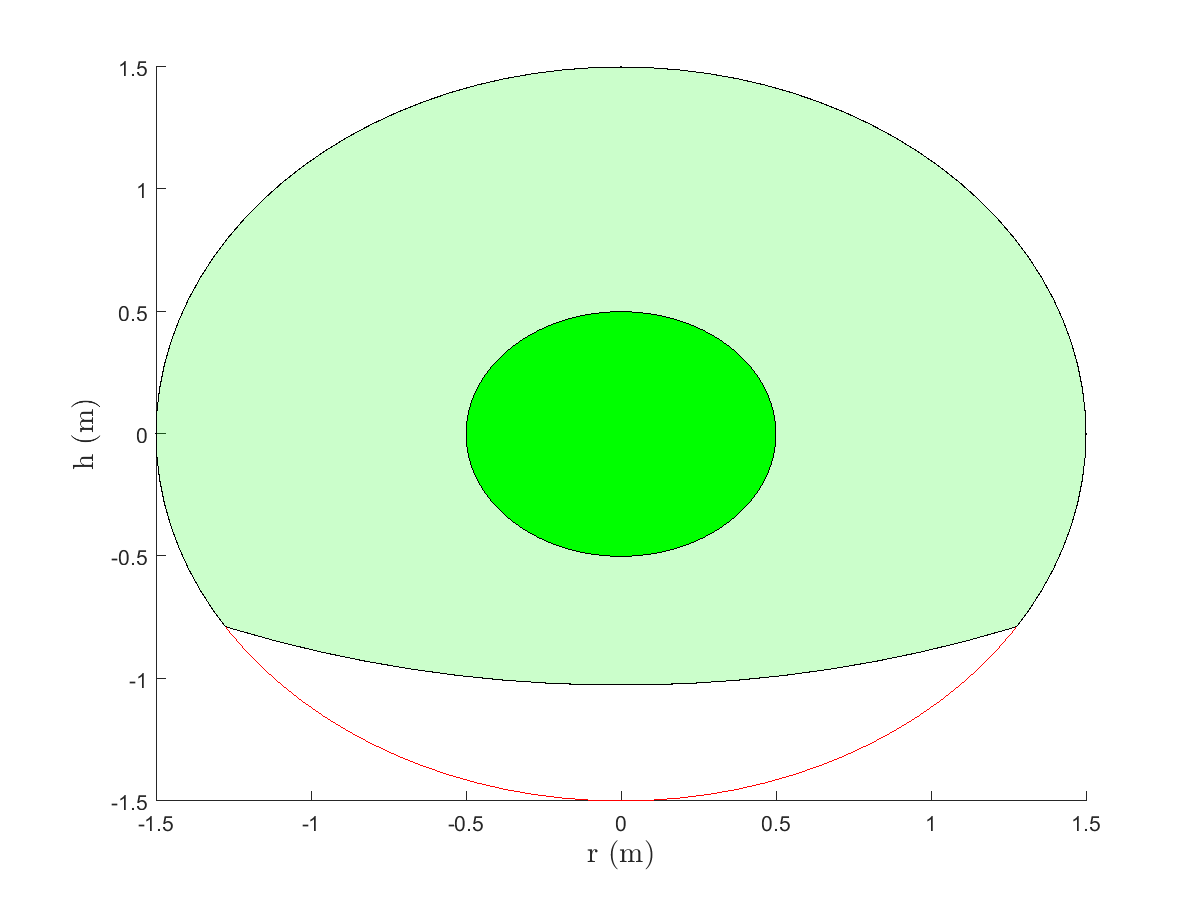}}  \\
{\includegraphics[scale=0.21,trim = 0cm 0cm 0cm 0cm,clip=true]{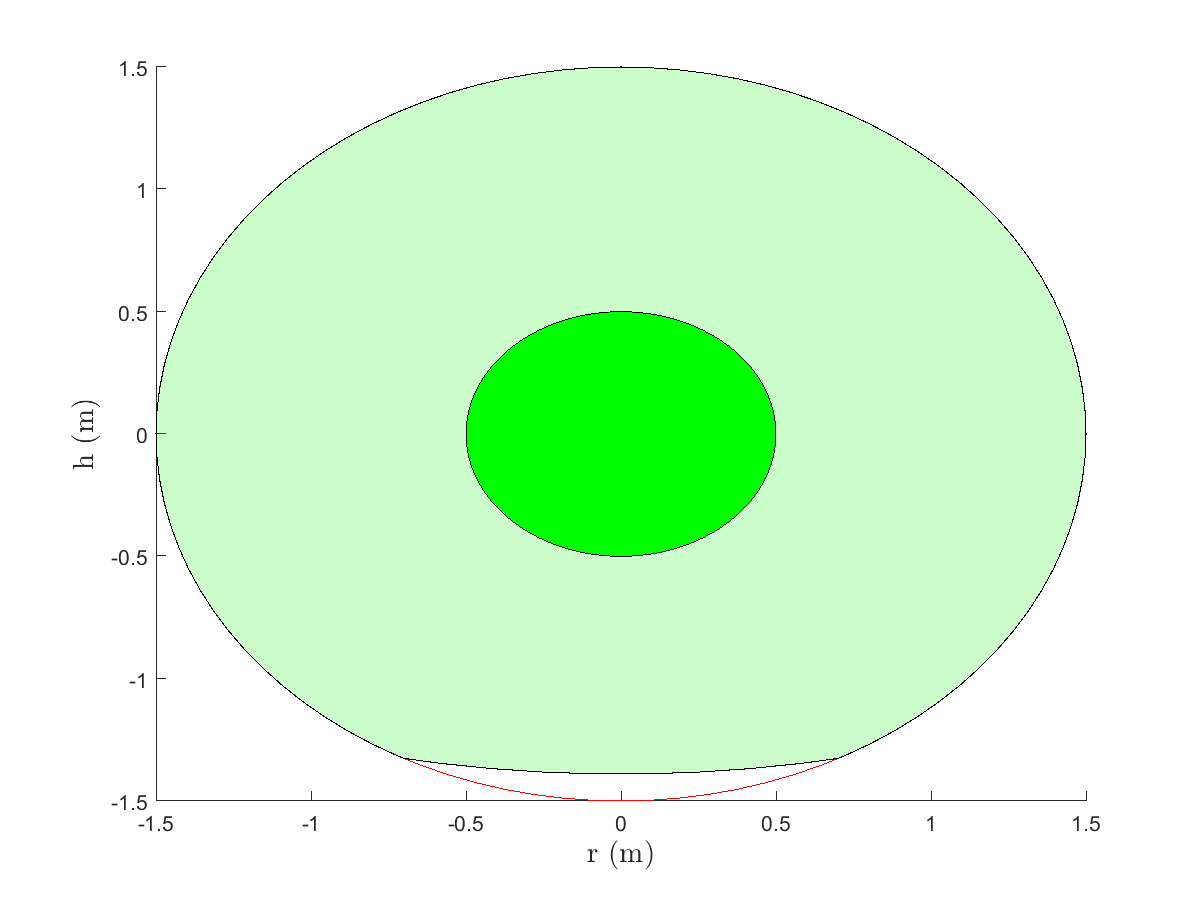}}  &
{\includegraphics[scale=0.21,trim = 0cm 0cm 0cm 0cm,clip=true]{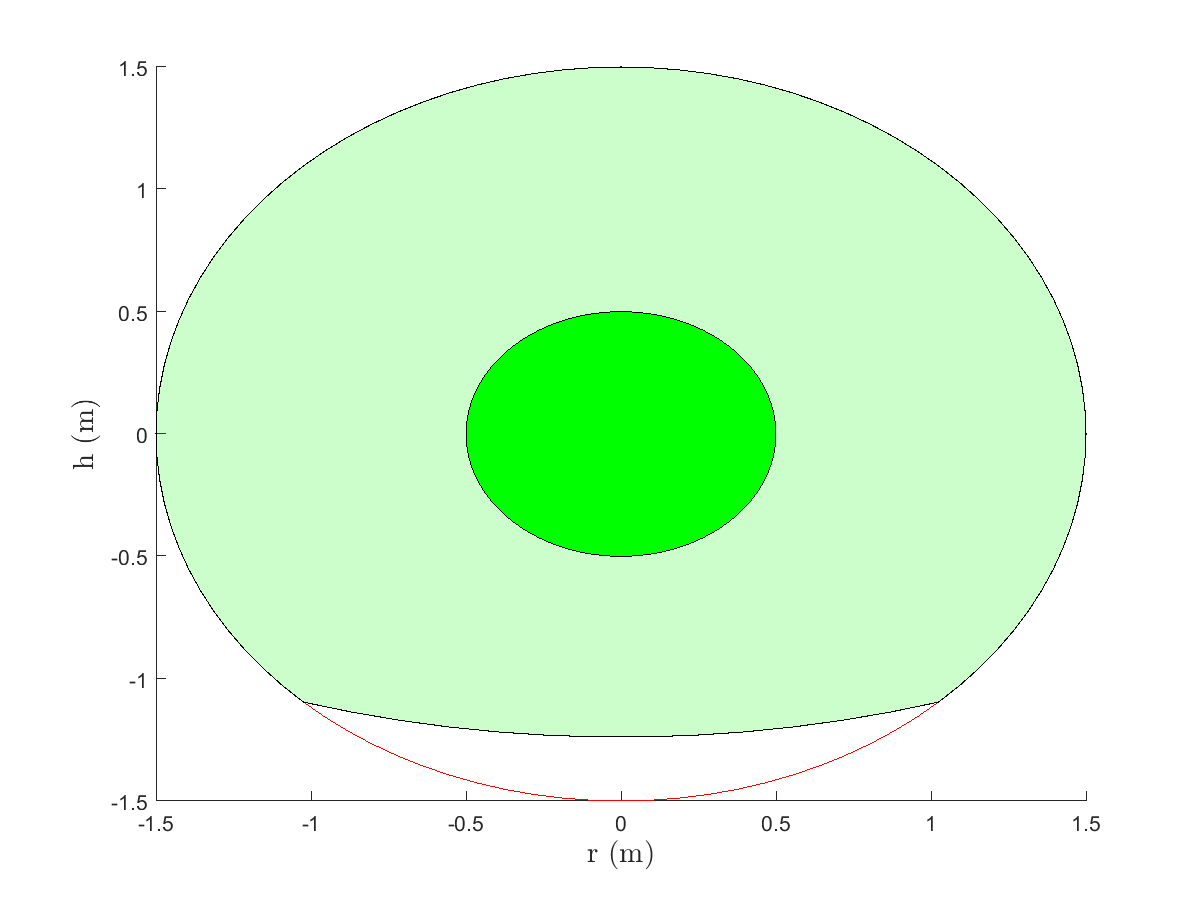}}  
\end{tabular}}
\vspace{-0pt}
\caption[]{Example crossarm region coverage for $\mathbf{p}_q(0.5,0,z)$, $\phi=\theta=\psi = 0$, $\mathbf{u}=\mathbf{0}$ with $z\in\{-2.5,-2.0,-1.5,-1.0\}$ (clockwise from upper left) and RealSense field of view $H_C,V_C,R_C$. The crossarm region $\partial \mathcal{C}$ (\textcolor{red}{$\square$}) is shown along with the sensor coverage of the crossarm region $\partial \mathcal{S}_{c}$ (\textcolor{green}{$\square$}) and sensor coverage of the pole region $\partial \mathcal{S}_{p}$  (\textcolor{green}{$\blacksquare$}).}
\vspace{-16pt}
\label{p2}
\end{figure}
To analyse the probability of detecting the pole, crossarm and wires using optical sensors, the coverage of the crossarm and detection disc is analysed at each $r_q \in (0.5,5.0)$ and $\Delta h \in (-4,4)$ with $y=0$. First, the bounding polygon $\partial \mathcal{S}_{xy}$ that defines the intersection points between the $xy$-pane at crossarm height and the spherical section defining the field of view is found. Second the bounding polygons that define the intersection between $\partial \mathcal{S}_{xy}$ and the crossarm $\partial \mathcal{S}_{c}$ and detection $\partial \mathcal{S}_{p}$ region is found. Third, a crossarm coverage area metric $A_s/A_c$ and pole area metric $A_s/A_p$ is found where $A_s$, $A_c$ and $A_p$ are the area of the bounding polygons defined by $\partial \mathcal{S}_c$, $\partial \mathcal{C}$ and $\partial \mathcal{S}_p$ respectively. Example sensor coverage results and parameters for a single location are shown in Fig.~\ref{p1}, with results for multiple locations shown in Fig.~\ref{p2} - \ref{r2}.

Using these results, a candidate protection can then be defined such that $r_z \geq 0.5$, $h_z\geq 4$ and $d_z \geq 2.2$ meters. The regions where the crossarm and wires are more likely to be detected given sensor constraints are $0.5 \leq r_q \leq 3.5$ and $-2 \leq \Delta h \leq 2$. The important point to note is the relative distance between the assumed infrastructure parameters and the protection zones. Altering the infrastructure parameters will change the protection zones accordingly, so knowledge of exact parameters allows a solid starting point to derive specific protection zones for each application or flight. 

Using the same set of coverage results, the sensor location on the platform can be determined. If a single forward facing sensor (non-gimbaled) is used with its optical axis is aligned with the platforms $x$ axis, the field of view is obviously limited and the possibility of missed detection increases at safety critical times during inspection. As the platform moves toward the pole, crossarm and wire detection becomes difficult as they leave the field of view. These effects are shown in Fig.~\ref{r1} and \ref{r2}, where the notion of tilting is captured by rotating the sensor downward in pitch. An obvious insight is to then use two optical sensors, whereby one is upward facing and the other is downward facing with overlap areas near the $x$ axis. The result is an increased probability of detection, moving inward, outwards or hovering near the infrastructure, alleviating the gimbals and providing some robustness to missed detections (i.e. one camera is required to detect the asset)

\noindent {{\textbf{Remark 1:} Using the above analysis: 
\begin{enumerate}[$\blacktriangleright$]
\item Two Intel RealSense cameras are mounted such that their optical axis are offset from the platform $x$ axis by $30^{\circ}$ and  $-30^{\circ}$. The field of view of each camera is $\{H_C,V_C\}$ is $-30^{\circ} \leq H_C \leq 30^{\circ}$ horizontally and  $-23^{\circ} \leq V_C \leq 23^{\circ}$ vertically. Combining the two cameras gives us a sensor with a vertical field of view of $\pm 53^{\circ}$ with a gap of $7^{\circ}$ in the middle.

\end{enumerate}
}
}

\noindent{\textbf{Remark 2:} The maximum permissible pitch angle is 20 degrees whilst the maximum permissible roll angle assumed to be 0 degrees. These constraints are based on the assumption that the platform will be oriented such that the platform's $x$ axis points toward the pole centre at all times. For the initial analysis presented here, the control input at each location is such that $\mathbf{u} = \mathbf{0}$ so $\phi=\theta=\psi = 0^{\circ}$ and $\mathbf{u} = \mathbf{0}$ so $\phi=0, \theta\pm 20^{\circ}, \psi= 0^{\circ}$ in Fig.\ref{r1} and \ref{r2} respectively.}

\noindent{\textbf{Remark 3:} The above analyse can be repeated for any sensor field of view, infrastructure parameters and platform motion. The results are to be used a guide to design the concept of operation and are can be modified for other platform constraints and refined after a practical assessment of the sensor performance.}

\subsection{Detection Algorithm/Approach}
The input to the pole detection algorithm consists of two point clouds captured using the RealSense cameras on-board, and the output is the the horizontal distance and the relative angles to the pole.

The detection process starts by combining the two clouds using the static rigid transform between the center of the UAV and the two cameras. Following that, the resulted combined cloud is passed through a noise reduction stage that deploy a voxel grid filter followed by a radius outlier removal. Radius outlier removal calculates
the number of neighbors within a given radius for each
point. Outliers are all points that have fewer than a
given number of neighbors

The pole detection algorithm works under the assumption that the pole is the longest vertical structure in the current view. Based on this assumption, the points in the filtered point cloud are used to fill a 2D histogram with a bin of size $25$cm $\times$ $25$cm. This is done by projecting the points into the $x-y$ plane. The center of the pole is the bin with the highest number of points count. Given the detected center of the pole in the reference frame of the UAV, the horizontal distance and the relative angles to the pole are calculated and returned to the assisted control system.

\begin{figure}[t!]
\centerline{
\begin{tabular}{c}
{\includegraphics[width=0.8\columnwidth]{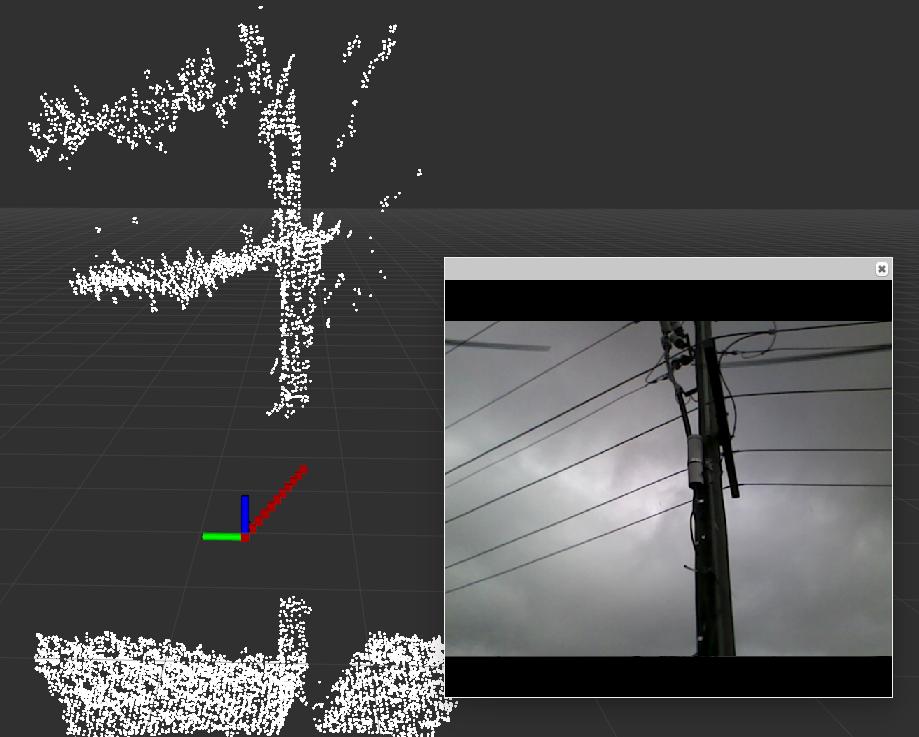}} 
\end{tabular}}
\vspace{-8pt}
\caption[]{The filtered cloud with the pole detected. The red dashed line is the vector between the UAV and the detected center of the pole.}
\vspace{-14pt}
\label{percep1}
\end{figure}

Fig.~\ref{percep1} shows the filtered cloud after voxelisation and outlier removal. It also shows a vector between the center of the UAV and pole based on the 2D histogram voting as descried above. The gap in the cloud is explained in section~\ref{sensorConfig}, Remark~1.

\section{Assisted Control} \label{cont}
\noindent The platform has 2 modes of operation or flight modes - Manual Mode and Inspect Mode. 

\subsection{Manual Mode} \label{mm}
\noindent The manual mode allows the platform to be controlled manually and is typically used pre and post inspection task. The manual mode can be instigated during inspection by the operator or in the case of missed infrastructure detections, and is typically used to start, pause (hover) or return from the inspection task. No assisted control is included such that
\begin{equation}
\mathbf{u} = \tilde{\mathbf{u}} = (\tilde{u}_x \,\tilde{u}_y \,\tilde{u}_z \,\tilde{u}_{\omega})
\end{equation}
Automation in manual mode is inherited from the onboard autopilot functionality which includes the ability to hover at a fixed location and heading when no control input is given. 

\subsection{Inspect Mode} \label{im}
\begin{figure}[t!]
\vspace{0pt}
\centerline{
\begin{tabular}{c}
{\includegraphics[scale=0.4,trim = 1cm 5cm 5cm 1.5cm,clip=true]{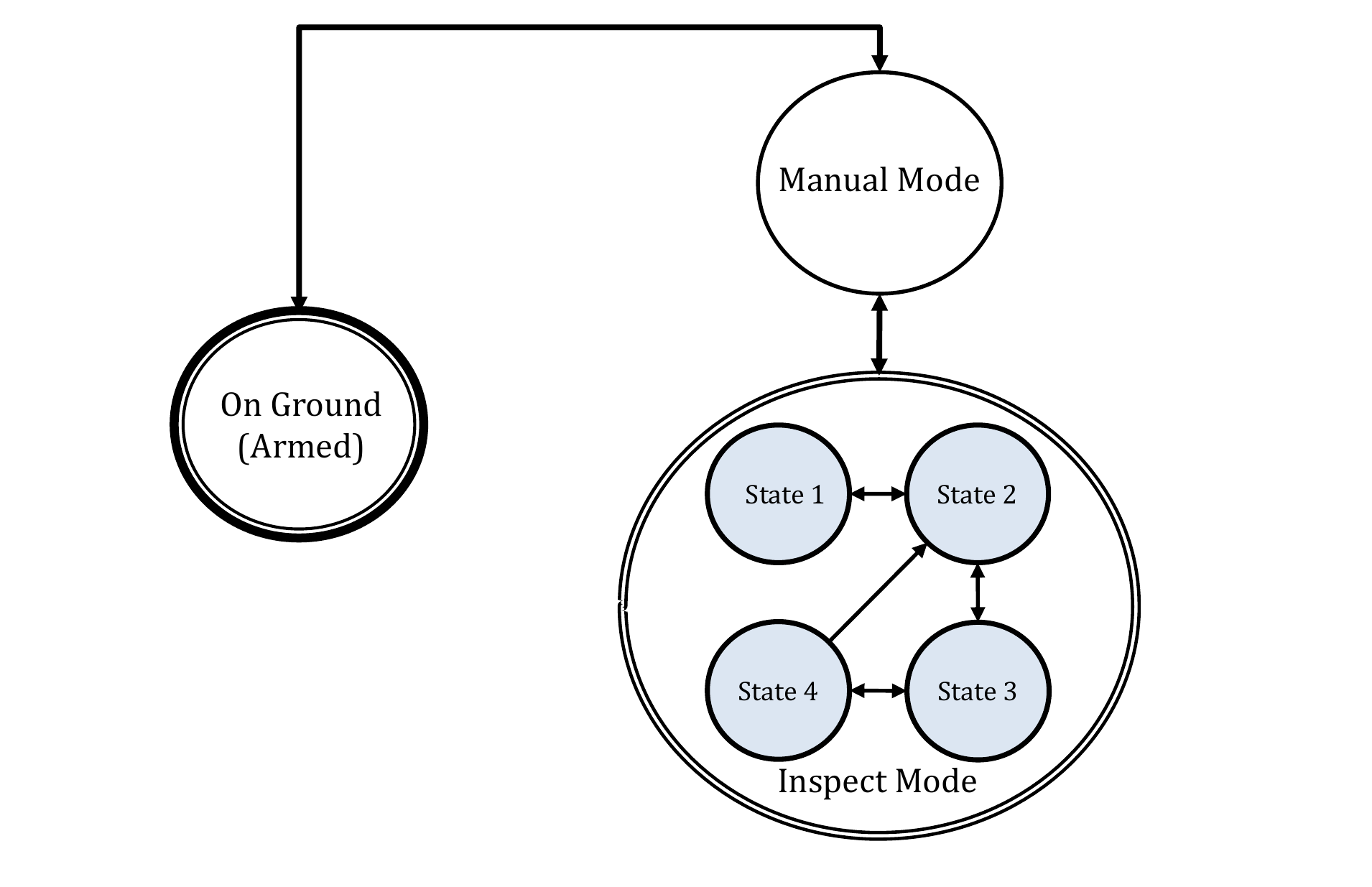}}
\end{tabular}}
\vspace{-8pt}
\caption[]{Flight mode transition architecture where one and two-way mode transitions are shown as well as the inspect sub-modes (\textcolor{blue}{$\blacksquare$}).}
\vspace{-14pt}
\label{d3}
\end{figure}

\noindent The inspect mode incorporates the safety features through assisted control that will alter manual control inputs that would place the platform inside the protected zones around infrastructure. Due to sensing constraints outlined earlier, the inspect mode operates in 4 states or sub-modes. Each state depends on where the platform is located with respect to the crossarm and wires and detection status. A description of each state and associated control input is provided below with the state transition topology depicted in Fig.~\ref{d3}. In this paper, we only consider control in the horizontal $xy$-plane (3DOF) such that $u_z = \tilde{u}_z$ in all states.

\noindent\rule{8.6cm}{0.4pt}\\
\noindent \textbf{State 1}
\noindent The inspection mode has been initiated but infrastructure has not yet been detected so assisted control cannot be applied. The platform is essentially in manual mode such that $\mathbf{u} = (\tilde{u}_x \,\tilde{u}_y\,\tilde{u}_z\,\tilde{u}_{\omega})$, but can automatically switch to states 2-4 pending relative geometry and sensing performance. This is not the same as in manual mode, and can be seen as an arming of the inspect mode. This state is the only state that can be entered from manual mode. If the inspection is paused this state is re-entered when inspection resumes, but may (almost) immediately switch to states 2-4 subject to the relative geometry and requested control input.
 
\noindent\rule{8.6cm}{0.4pt}\\
\noindent \textbf{State 2}
The platform has detected the infrastructure and the safety system attempts to orientate the platform such that the platforms $x$ axis always points toward the pole within a specific deadband via proportional-derivative yaw control such that:\vspace{-8pt}

\footnotesize
\begin{equation} u_{\omega} =
\begin{cases} 
K_p \psi(t) + K_d (\psi(t)-\psi(t-1))/\delta t \quad  &|\psi(t)-\psi(t-1)| > \tau \\
\tilde{u}_{\omega} \quad  &|\psi(t)-\psi(t-1)| \leq \tau
\end{cases}
\end{equation}
\normalsize where $K_p$ and $K_d$ are the proportional and derivative gain terms respectively and $\psi$ is the relative heading angle from the quad $x$-axis and the pole centre. The controller allows manual yaw corrections based on the value of $\tau \in \mathbb{R}^+$ where tau is the angle between $r_q$ and $x$ axis of the platform. The requested control input is such that there is no collision threat allowing 
\begin{equation}
\mathbf{u} = (\tilde{u}_x \,\tilde{u}_y\,\tilde{u}_z\,u_{\omega})
\end{equation}

\noindent\rule{8.6cm}{0.5pt}\\
\noindent \textbf{State 3}
\noindent The platform has detected the infrastructure and the safety system attempts to re-orientate the platform in the same manner as state 2. The requested control input is such that there is a potential collision threat. The safety system ensures that the platform cannot enter into the protected zone surrounding the pole such that $r_q > r_z$. The operator remains in control of the inspection task, with the requested control modified such that:
\begin{eqnarray}
u_x &=& 
\begin{cases}
\tilde{u}_x(1 - 1/e^{|r_q - r_z|})v,&\quad \psi_v \leq \eta \\
\tilde{u}_x,&\quad |\psi_v| > \eta 
\end{cases} \\
u_y &= &
\begin{cases}
\tilde{u}_y(1 - 1/e^{|r_q - r_z|})v,&\quad \psi_v \leq \eta\\
\tilde{u}_y,&\quad |\psi_v| > \eta  
\end{cases} \\
\end{eqnarray} 
where $\psi_v$ is the angle between the requested lateral velocity vector and the pole centre measured from the quad centre of mass. The angle $\eta$ defines the angle between the vector tangential to the inner collision boundary $r_z$ and the pole, measured from the platform.

\noindent\rule{8.6cm}{0.4pt}\\
\noindent \textbf{State 4}
\noindent The platform has detected the infrastructure and the safety system attempts to re-orientate the platform in the same manner as state 2. The position of the platform is such that there is an immediate collision threat, as the inner boundary has been breached. The safety system ensures that the platform exits the protected zone such that $r_q > r_z$, Operator input is modified such that:
\begin{eqnarray}
u_x &=& 
\begin{cases}
-\tilde{u}_x(1 - 1/e^{|r_q - r_z|})v,&\quad \psi_v \leq \eta \\
\tilde{u}_x,&\quad |\psi_v| > \eta 
\end{cases} \\
u_y &= &
\begin{cases}
-\tilde{u}_y(1 - 1/e^{|r_q - r_z|})v,&\quad \psi_v \leq \eta\\
\tilde{u}_y,&\quad |\psi_v| > \eta  
\end{cases} \\
\end{eqnarray} 
forcing the platform outside the inner boundary regardless of the controller input. Importantly, this control design also guards against collision in the event the operator input is incorrect due to disorientation, panic or otherwise.

\section{Infrastructure Inspection} \label{sec:exp}
To create a working platform capable of conducting real inspection tasks, the perception system is coupled with the assisted control design in a multi-loop control architecture (nested) operating at different frequencies. The pilot input defines the reference control input which, depending on the state, is modified through the high-level controller and then sent as a reference for the low-level attitude controller (Pixhawk). The high level controller runs at $f_H=1$Hz and the low-level controller runs at $f_L=400$Hz. This same control architecture/topology is used in both simulation and on the real platform. Both simulated and real experiments are conducted to evaluate the safety system for use in live inspection tasks.

\subsection{Simulated Results}
\begin{figure*}[t]
\vspace{0pt}
\centerline{
\begin{tabular}{cc}
{\includegraphics[scale=0.4,trim = 3cm 9cm 3cm 9cm,clip=true]{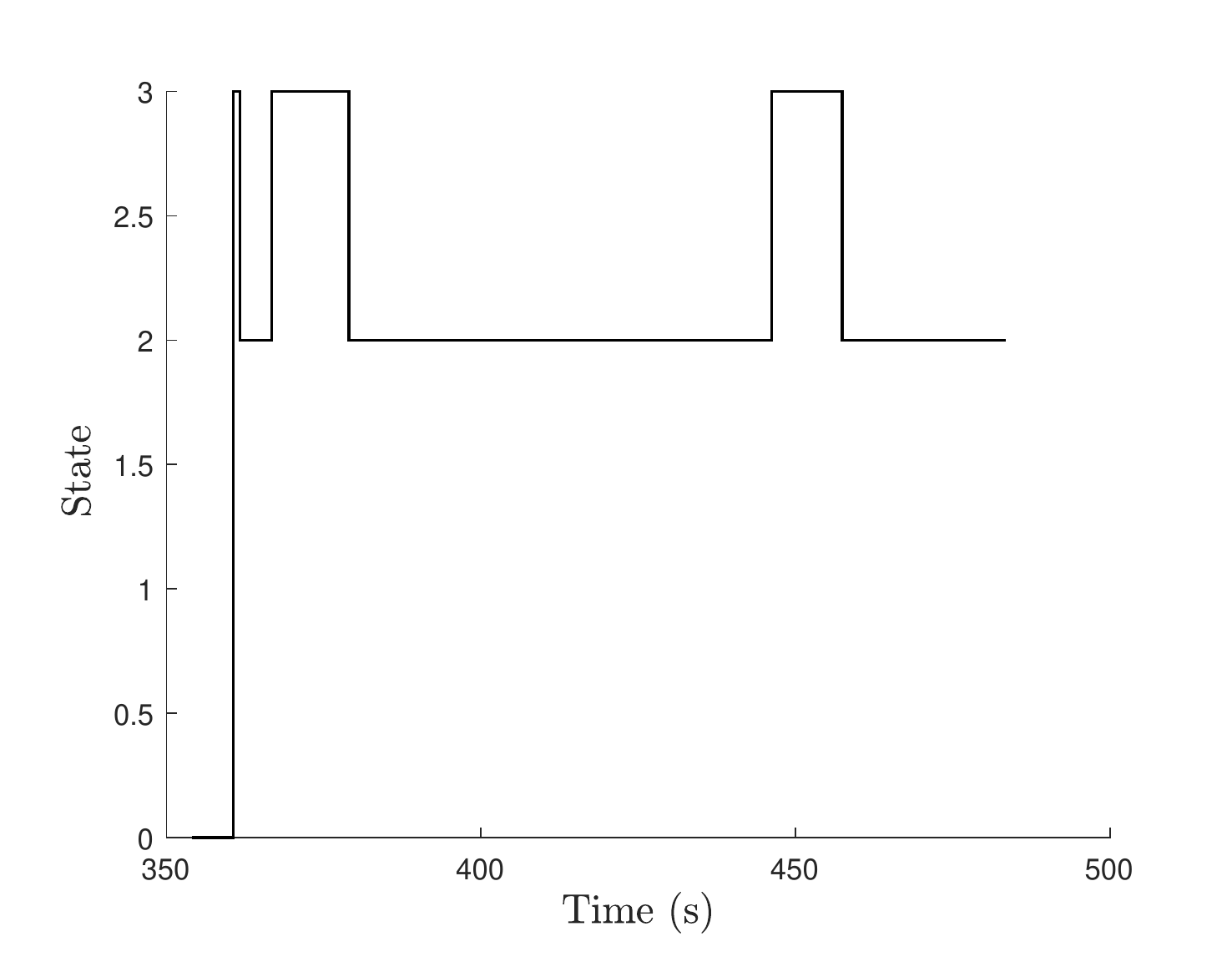}} &
{\includegraphics[scale=0.4,trim = 3cm 9cm 3cm 9cm,clip=true]{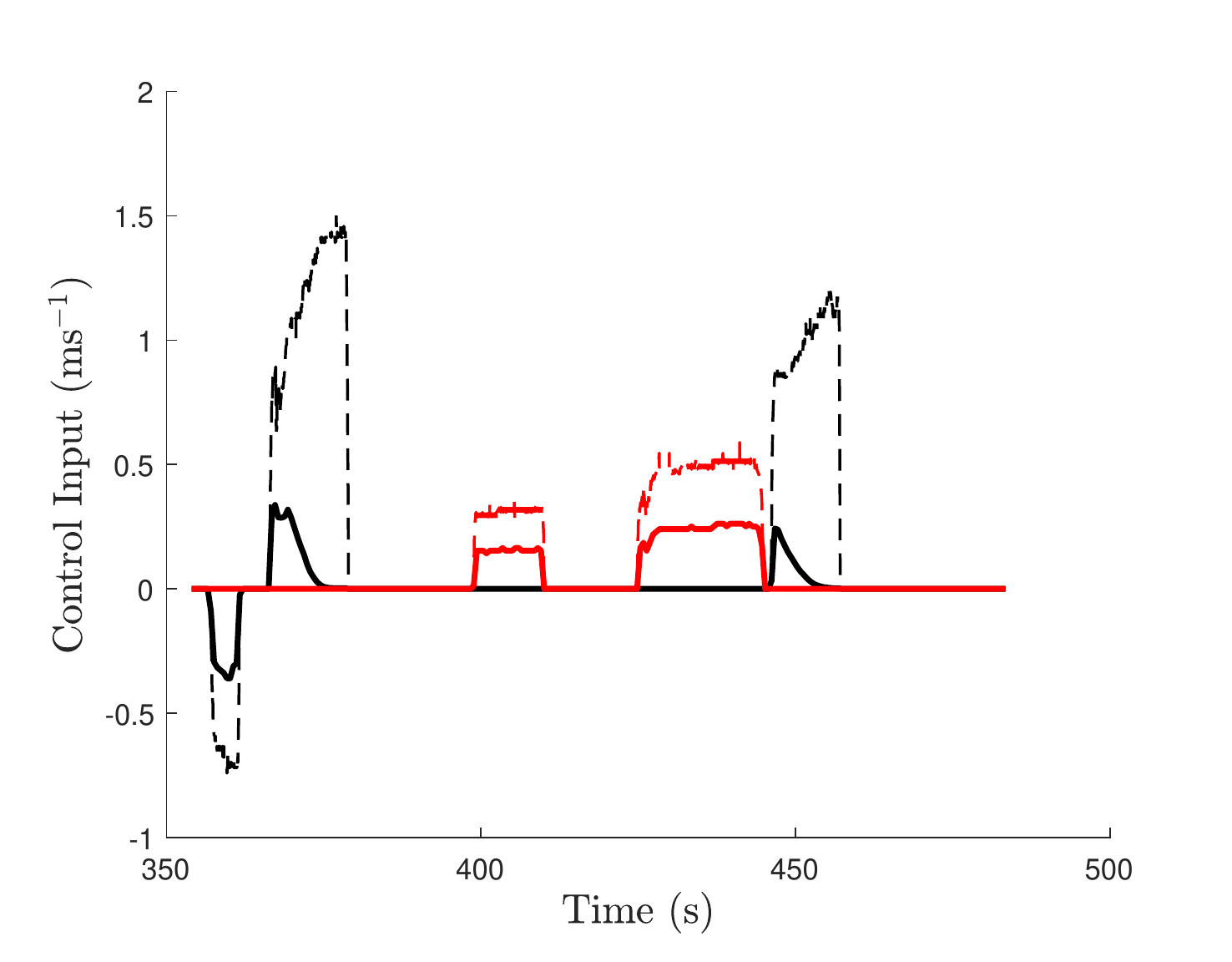}} \\
\footnotesize{(a) System state }& \footnotesize{(b) Horizontal ($xy$) control $\tilde{u}_x$ (\textcolor{black}{--}), $u_x$ (\textcolor{black}{-}) and $\tilde{u}_y$ (\textcolor{red}{--}), $u_y$ (\textcolor{red}{-})} \\
{\includegraphics[scale=0.4,trim = 3cm 9cm 3cm 9cm,clip=true]{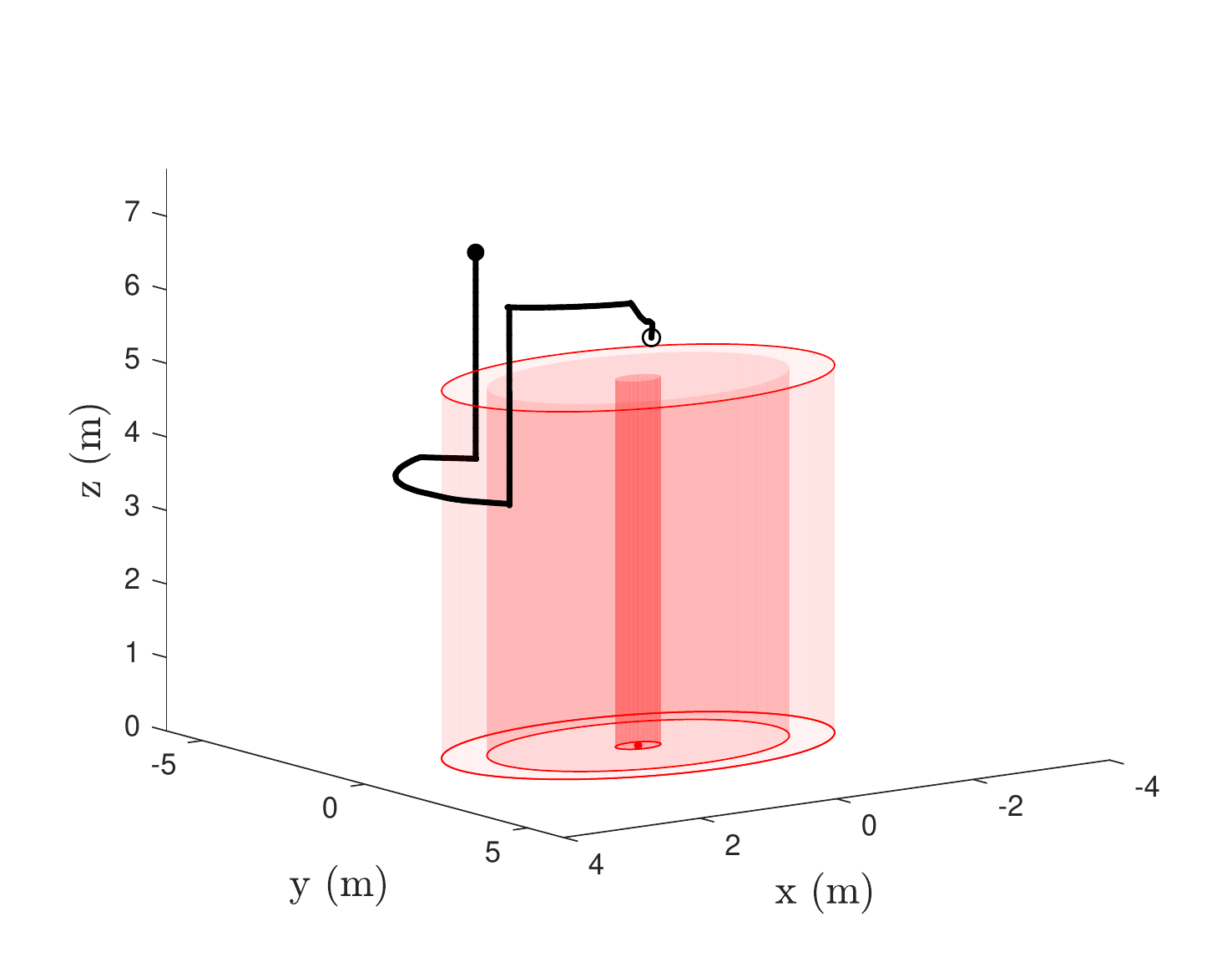}} &
{\includegraphics[scale=0.5,trim = 3cm 9cm 3cm 9cm,clip=true]{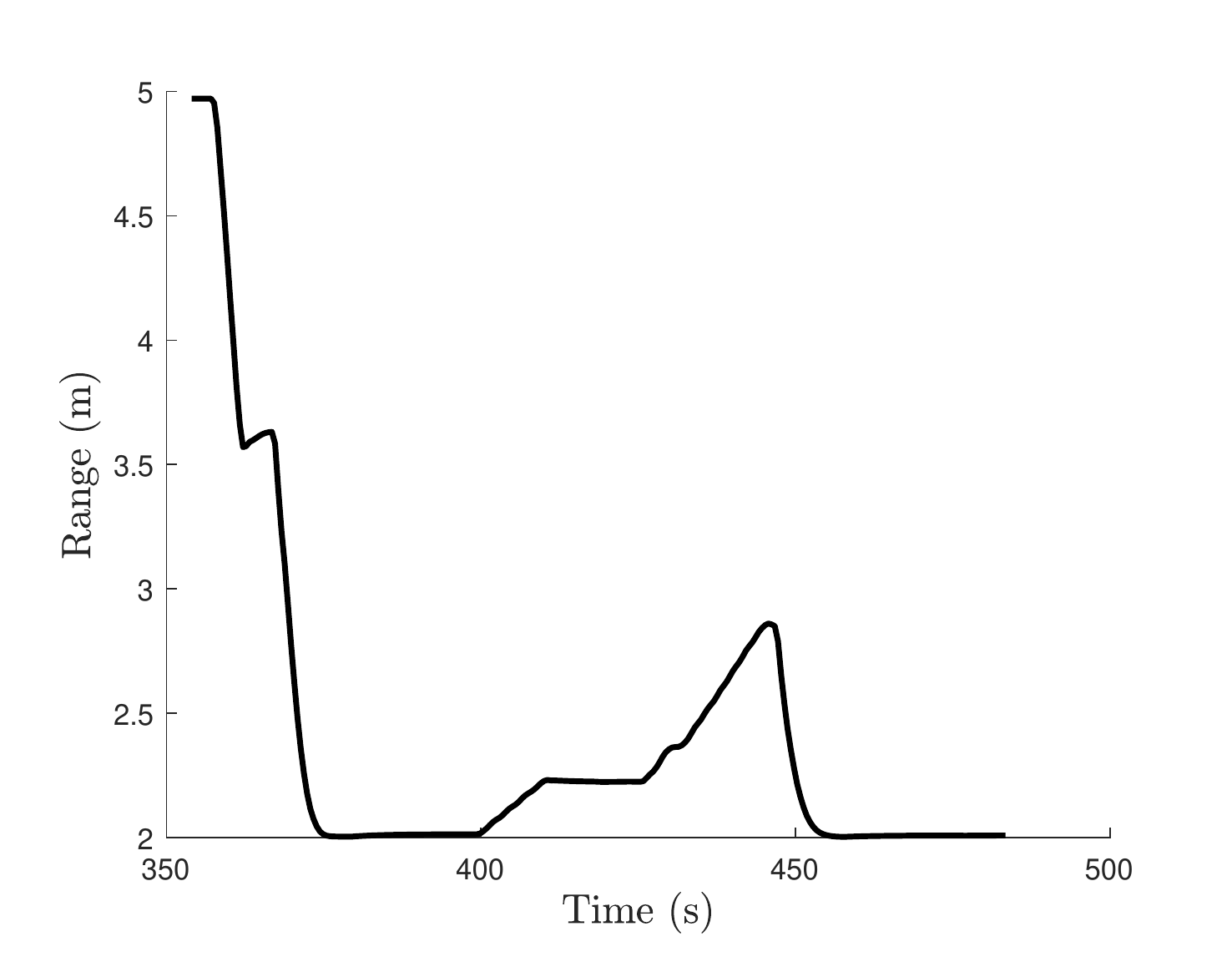}} \\
\footnotesize{(c) Flight path (\textcolor{black}{-}), pole (\textcolor{red}{$\bullet$}) and protection zone (\textcolor{red}{$\circ$})} & \footnotesize{(d) Range ($r$)} 
\end{tabular}}
\vspace{-4pt}
\caption[]{Simulated inspection flight with $\mathbf{p}_c(0, 0, 5)$, $r_c = 1.5$, $r_z=2.0$. Notice the platform state transition when operating near and away from the pole and the automatic control attenuation ($t\approx 375$ and $t\approx 440$) if requesting an unsafe input when operating near the pole.}
\vspace{-16pt}
\label{sim}
\end{figure*}

Simulations are conducted using a simple quadrotor model (\textit{hector\_quadrotor} \cite{2012simpar_meyer}) running in ROS and using Gazebo with an appropriate pole model to provide the feedback for the assisted control. The manual control uses hardware in the loop via a real RC transmitter. A simulation is conducted whereby the platform initially moves toward to pole, descends, orbits the infrastructure, moves toward the pole than ascends. The platform state, control input, flight path and relative distance (range) to the pole are depicted in Fig.~\ref{sim}(a)-(d) respectively.

Unlike the real experiment that we will detail later, the perception system in the simulation case is essentially perfect, so missed detections are not observed. Thresholds are placed on the relative yaw orientation and distance to the pole such that the assisted control can become active during flight. The simulation results demonstrate two main features. First, the assisted control successfully alters the control and consequently stops the operator from moving beyond the inner protection zone boundary. This is indicated by the platform moving into state 3 and the control subsequently converging to zero, even if the operator attempts to move further toward the pole. Second the yaw control applied in state 1 through to state 3 ensures that the platform points toward the pole, resulting in the orbit-like behaviour despite no control input in the $x$ direction ($t\approx400$s until $t\approx450$s).

\subsection{Real Experimental Results}
\begin{figure*}[t]
\vspace{0pt}
\centerline{
\begin{tabular}{cc}
{\includegraphics[scale=0.4,trim = 3cm 9cm 3cm 9cm,clip=true]{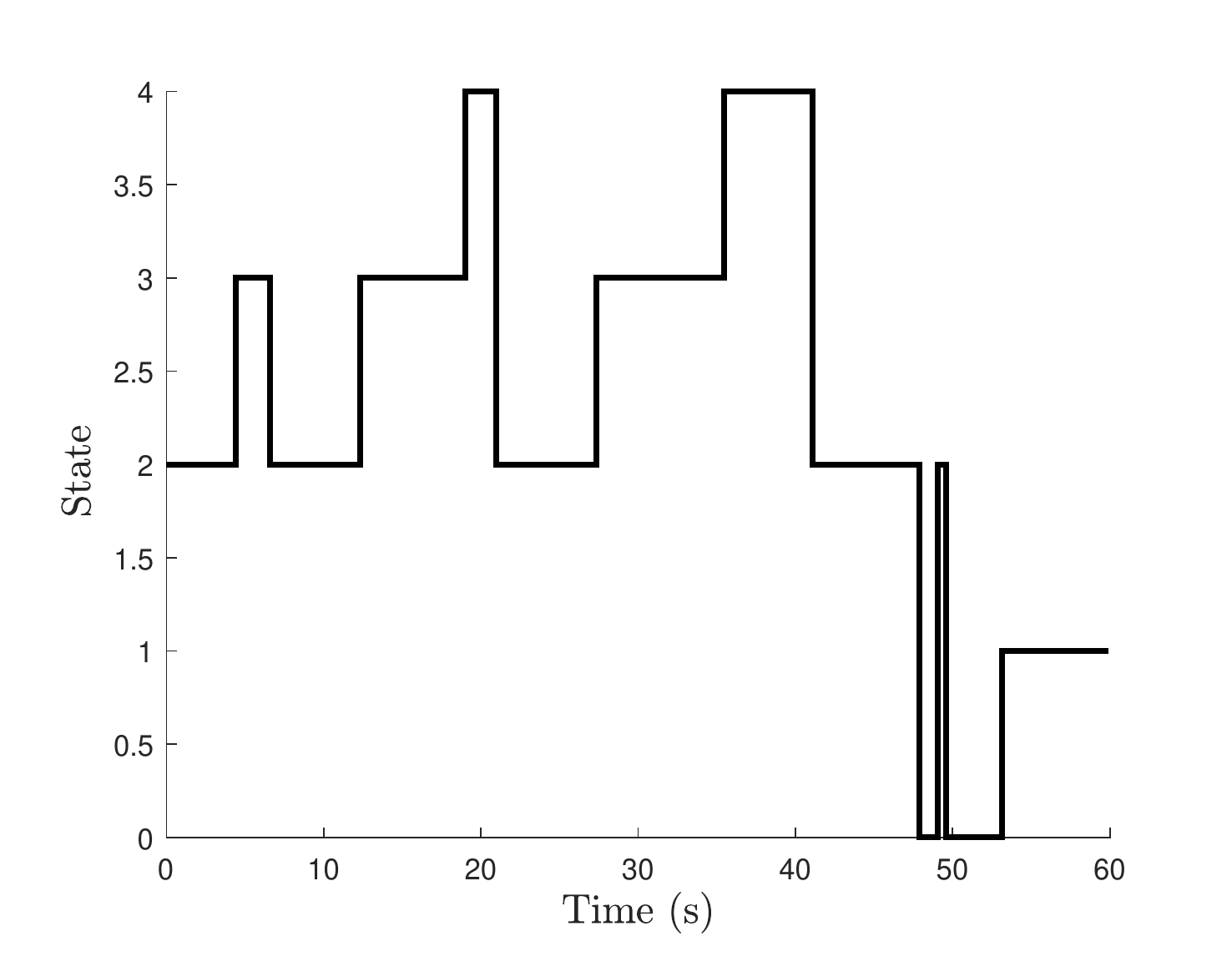}} &
{\includegraphics[scale=0.4,trim = 3cm 9cm 3cm 9cm,clip=true]{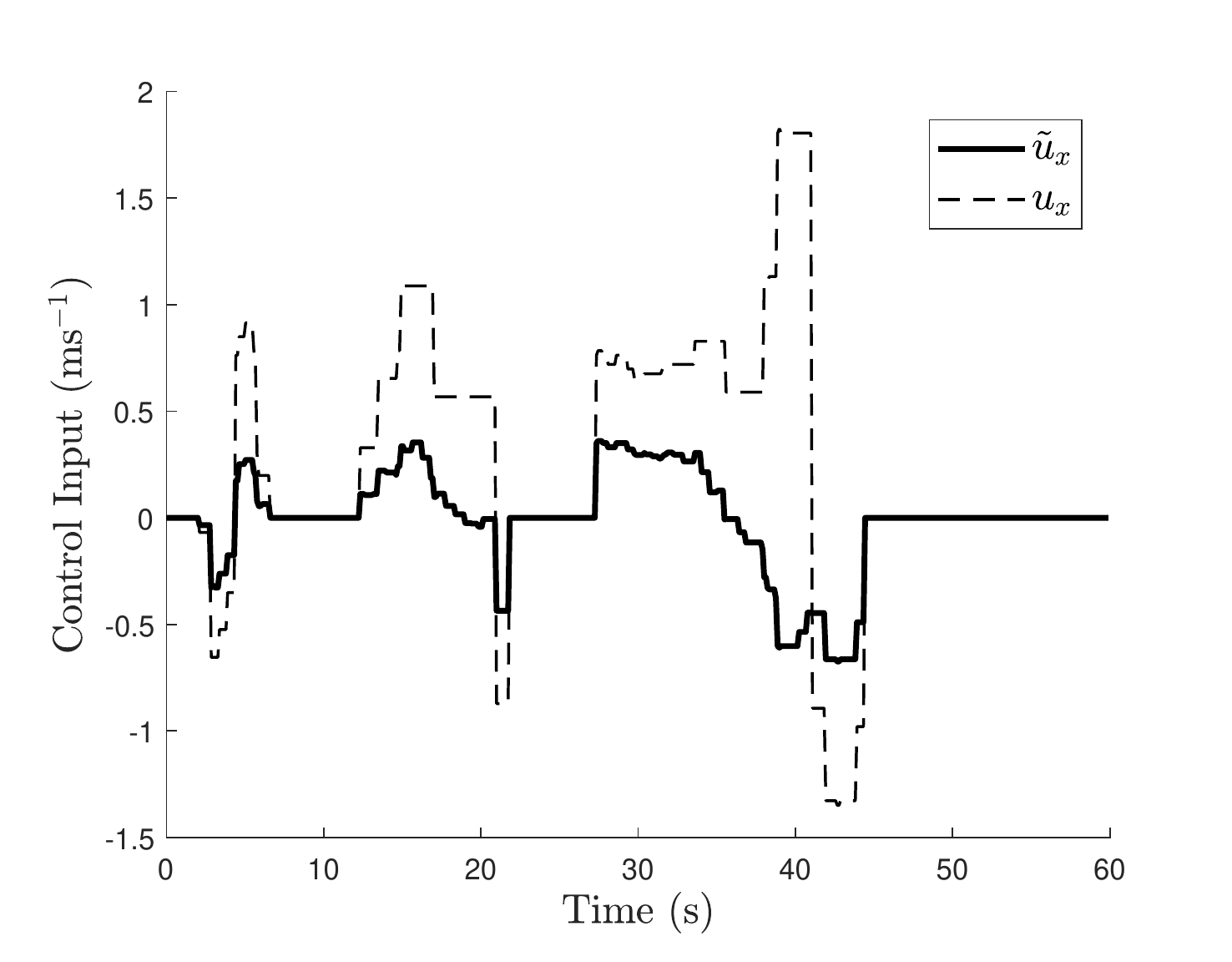}} \\
\footnotesize{(a) System state}  & \footnotesize{(b) Longitudinal ($x$) control $\tilde{u}_x$ (\textcolor{black}{--}), $u_x$ (\textcolor{black}{-}) and $\tilde{u}_y$ (\textcolor{red}{--}), $u_y$ (\textcolor{red}{-})} \\
{\includegraphics[scale=0.4,trim = 3cm 9cm 3cm 9cm,clip=true]{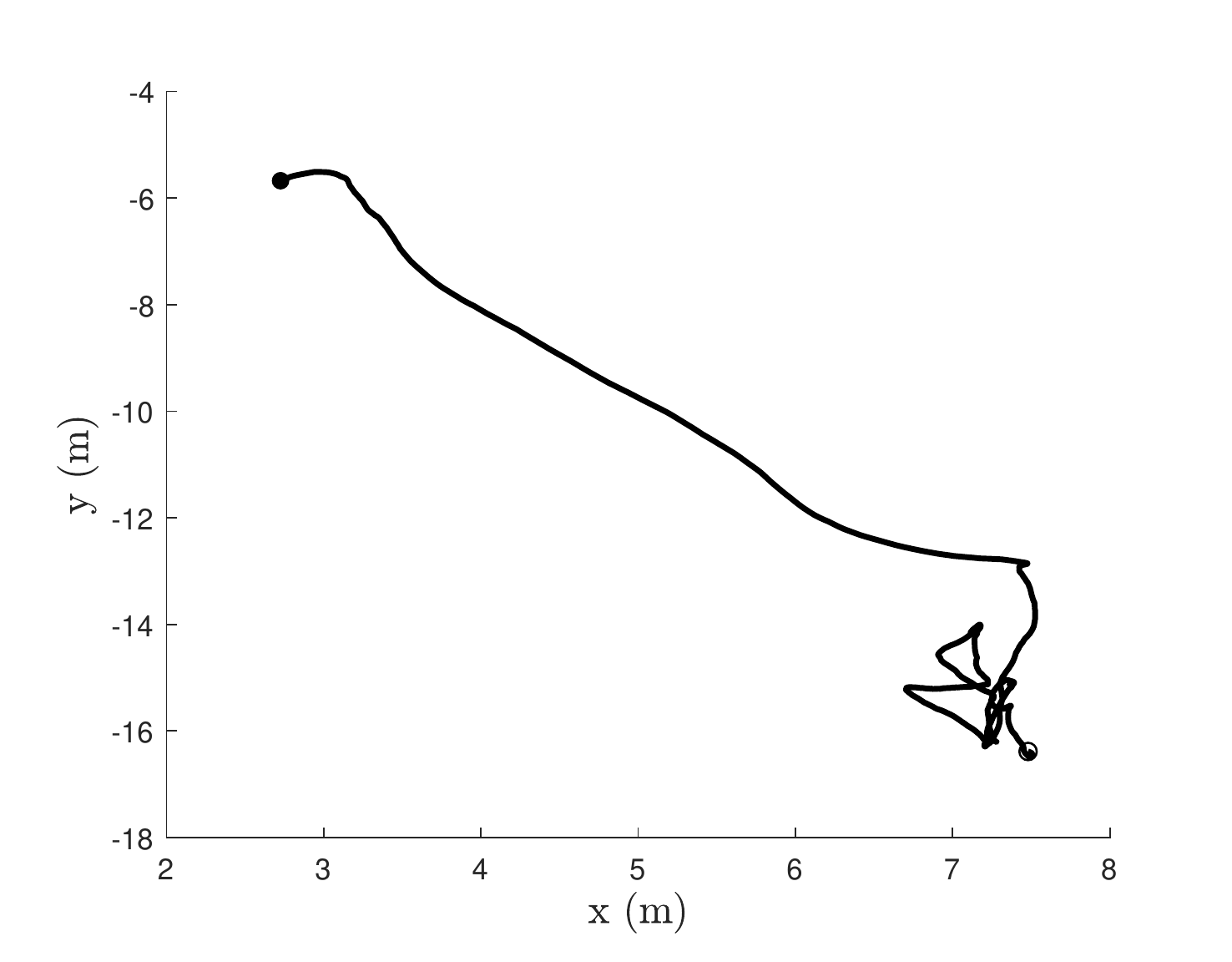}} &
{\includegraphics[scale=0.4,trim = 3cm 9cm 3cm 9cm,clip=true]{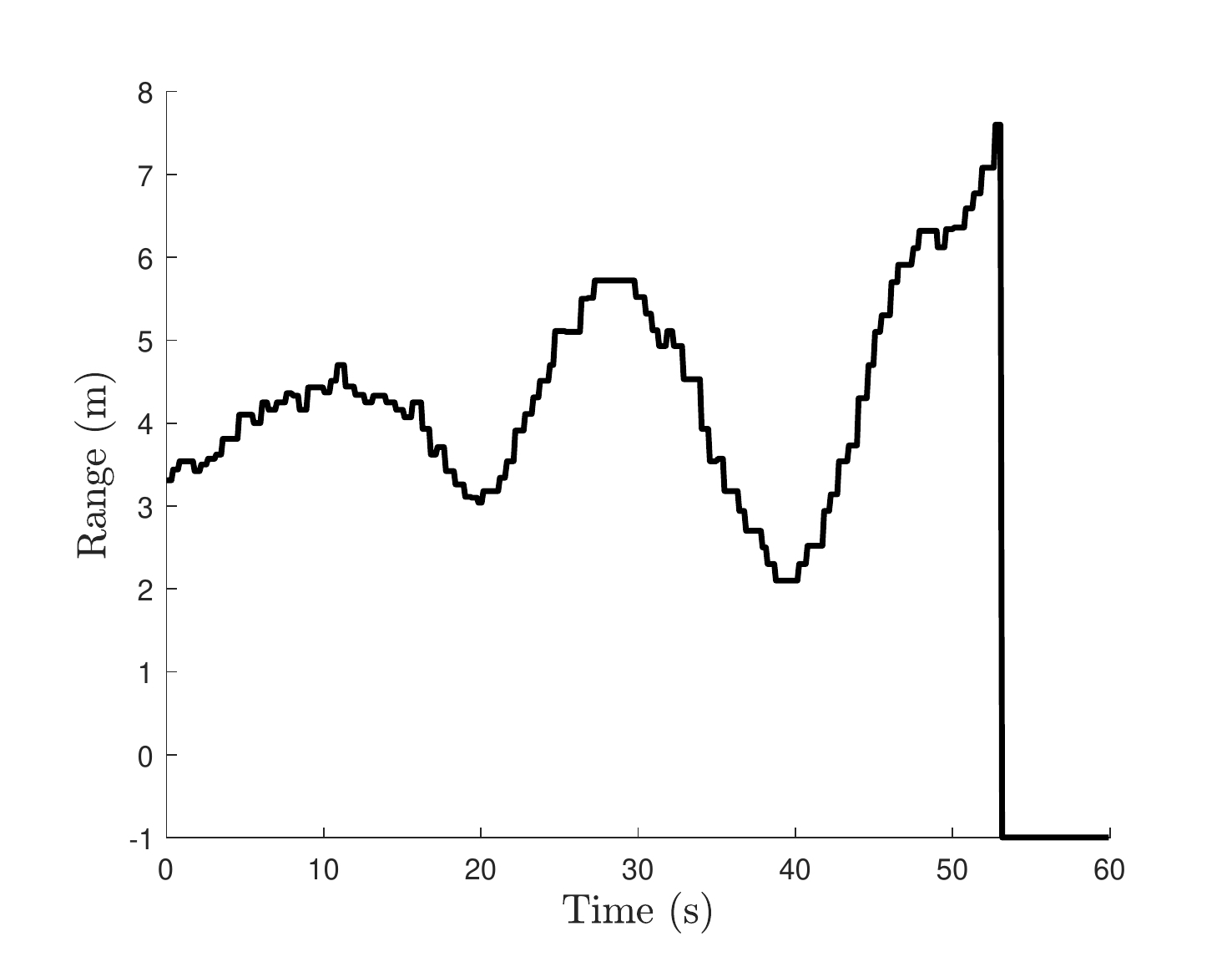}} \\
\footnotesize{(c) Flight path (\textcolor{black}{-}) with start ($\circ$)} and end ($\bullet$) points & \footnotesize{(d) Range ($r$)} 
\end{tabular}}
\vspace{-2pt}
\caption[]{Real infrastructure inspection flight with $r_c = 1.1$, $r_z=3$. Notice the platform state transition to state 4 ($t\approx 20$) and ($t\approx 40$) and the subsequent control polarity reversal to move the platform away from the infrastructure. }
\vspace{-16pt}
\label{exp}
\end{figure*}

Flight trials are conducted using real (de-energised) electrical infrastructure used for training purposes in the power industry. All perception and control is accomplished onboard (via an Intel NUC), with an offboard laptop required for setup and two RC controllers used for safety. The master RC transmitter is always active and used to switch between manual and inspect modes. The second transmitter is used during inspect mode and is directly linked to the perception and assisted control systems. A flight is conducted whereby the platform starts near the infrastructure before attempting to move as close as possible on two consecutive attempts before retreating away from the asset. The platform state, control input, flight path and relative distance (range) to the pole are depicted in Fig.~\ref{exp}(a)-(d) respectively.

As the platform is close to the infrastructure, the perception system immediately detects the pole and initiates automatic yaw control to re-orient the platform toward the pole. The experimental results support the observation from the simulated results, but also demonstrate a key safety feature. Due to missed detections and measurement error, the platform moves beyond the protection zone boundary on two occasions as the operator attempts to inspect the infrastructure ($t\approx 20$ and $t\approx40$). The assisted control responds accordingly, rejecting the operator input and pushing the platform back outside the protection zone. This is indicated by the platform moving into state 4 and with the control input being the opposite polarity to that requested until the operator no longer attempts to move toward the pole. At this time, the control polarity is retained by attenuated accordingly to ensure non-aggressive behaviour.

\textit{\textbf{Remark:} The negative range value in Fig.~\ref{exp}(d) at $t\approx52$s occurs when the pole moves out of detection range (state 1), and the platform is essentially in manual mode.} 

Table~\ref{controltable} lists the parameters of the system for both the simulation and the real experiment case.
\begin{table}[p]
\caption{System Parameters}
\centering
\vspace{-5pt}
\begin{tabular}{ c  c  c}  
  \toprule  
  $^{\dag}$Parameter (units) & Simulated    & Experimental  \\ 
  \midrule 
  $r_z$ (s) 			& 40 					& 8 \\
  $r_c$ (s) 			& 0.1 					& 0.05\\   
  $r_p$ (s) 			& 0.1 					& 0.05\\                     
  $K_p$ (-) 			& 3 					& 3\\
  $K_d$ 	(-)			& 4 					& 4\\
  $m$ (kg)				& 1  					& 4.62  \\
  $\mathbf{p}_p$ (m) & (-2,1,-3)	& -\\
  \bottomrule
\end{tabular}
\vspace{-18pt}
\label{controltable} 
\end{table}

\section{Conclusions} \label{conc}
This paper presented a full theoretical analysis of expected sensor performance and constrained platform behaviour for the task of assisted control of a small multirotor platform when performing an aerial inspection of fixed energy infrastructure. We show how the theoretical analysis inform the design of the sensor rig and the choice of the optical sensor placement.  Using the measurement for the optical sensors on-board, the relative position between the platform and the asset (the electrical pole in our case) is estimated and used to enable the human operator inputs to be autonomously adjusted to ensure safe separation. 

The development time required from acquisition to implementation and then successful results was less than 3 weeks. This is due to the appropriate modelling and simulation. No incidents occurred during flight, including no control failures, component failures or inadequate sensing highlighting the importance of the theoretical analysis during developmental stages.

\section*{Acknowledgement}
\vspace{-0pt}
\noindent This work has been supported by the Cooperative Research Centre for Spatial Information (CRC-SI), whose activities are funded by the Business Cooperative Research Centres Programme.\\
This work is part of the joint CRC-SI, Ergon and QUT project \textit{Aerial Robotics for Close Proximity Infrastructure Inspection - Project Number 2.24.}

\bibliographystyle{IEEEtran}
\bibliography{refs.bib}

\begin{thebibliography}{1}
\providecommand{\url}[1]{#1}
\csname url@rmstyle\endcsname
\providecommand{\newblock}{\relax}
\providecommand{\bibinfo}[2]{#2}
\providecommand\BIBentrySTDinterwordspacing{\spaceskip=0pt\relax}
\providecommand\BIBentryALTinterwordstretchfactor{4}
\providecommand\BIBentryALTinterwordspacing{\spaceskip=\fontdimen2\font plus
\BIBentryALTinterwordstretchfactor\fontdimen3\font minus
  \fontdimen4\font\relax}
\providecommand\BIBforeignlanguage[2]{{%
\expandafter\ifx\csname l@#1\endcsname\relax
\typeout{** WARNING: IEEEtran.bst: No hyphenation pattern has been}%
\typeout{** loaded for the language `#1'. Using the pattern for}%
\typeout{** the default language instead.}%
\else
\language=\csname l@#1\endcsname
\fi
#2}}

\bibitem{Mth2015VisionAC}
K.~M{\'a}th{\'e} and L.~Busoniu, ``Vision and control for uavs: A survey of
  general methods and of inexpensive platforms for infrastructure inspection,''
  in \emph{Sensors}, 2015.

\bibitem{Mth2016VisionbasedCO}
K.~M{\'a}th{\'e}, L.~Busoniu, L.~Barab{\'a}s, C.-I. Iuga, L.~Miclea, and
  J.~Braband, ``Vision-based control of a quadrotor for an object inspection
  scenario,'' \emph{2016 International Conference on Unmanned Aircraft Systems
  (ICUAS)}, pp. 849--857, 2016.

\bibitem{Stokkeland2015AutonomousVN}
M.~Stokkeland, K.~Klausen, and T.~A. Johansen, ``Autonomous visual navigation
  of unmanned aerial vehicle for wind turbine inspection,'' \emph{2015
  International Conference on Unmanned Aircraft Systems (ICUAS)}, pp.
  998--1007, 2015.

\bibitem{Araar2014VisualSO}
O.~Araar and N.~Aouf, ``Visual servoing of a quadrotor uav for autonomous power
  lines inspection,'' \emph{22nd Mediterranean Conference on Control and
  Automation}, pp. 1418--1424, 2014.

\bibitem{Lam2009ArtificialFF}
T.~M. Lam, H.~W. Boschloo, M.~Mulder, and R.~van Paassen, ``Artificial force
  field for haptic feedback in uav teleoperation,'' \emph{IEEE Transactions on
  Systems, Man, and Cybernetics - Part A: Systems and Humans}, vol.~39, pp.
  1316--1330, 2009.

\bibitem{Brandt2010HapticCA}
A.~M. Brandt and M.~B. Colton, ``Haptic collision avoidance for a remotely
  operated quadrotor uav in indoor environments,'' \emph{2010 IEEE
  International Conference on Systems, Man and Cybernetics}, pp. 2724--2731,
  2010.

\bibitem{Hou2013RepresentationOV}
X.~Hou, R.~E. Mahony, and F.~Schill, ``Representation of vehicle dynamics in
  haptic teleoperation of aerial robots,'' \emph{2013 IEEE International
  Conference on Robotics and Automation}, pp. 1485--1491, 2013.

\bibitem{Sa2015InspectionOP}
I.~Sa, S.~Hrabar, and P.~I. Corke, ``Inspection of pole-like structures using a
  visual-inertial aided vtol platform with shared autonomy,'' in
  \emph{Sensors}, 2015.

\bibitem{2012simpar_meyer}
J.~Meyer, A.~Sendobry, S.~Kohlbrecher, U.~Klingauf, and O.~von Stryk,
  ``Comprehensive simulation of quadrotor uavs using ros and gazebo,'' in
  \emph{3rd Int. Conf. on Simulation, Modeling and Programming for Autonomous
  Robots (SIMPAR)}, 2012, p. to appear.

\end{thebibliography}

\end{document}